\pdfoutput=1

\documentclass[11pt]{article}

\usepackage[final]{acl}
\usepackage{amsmath,amssymb}
\usepackage{amsfonts}

\usepackage{times}
\usepackage{latexsym}
\usepackage[table]{xcolor} 
\usepackage{diagbox}
\usepackage[T1]{fontenc}

\usepackage[utf8]{inputenc}
\usepackage{pifont}
\usepackage{microtype}
\usepackage{pdflscape} 
\usepackage{multirow} 
\usepackage{inconsolata}

\usepackage{graphicx}
\usepackage{array}
\usepackage{colortbl}
\usepackage{booktabs}
\usepackage{caption}

\usepackage{tabularx}
\usepackage[table]{xcolor}
\usepackage{tcolorbox} 
\usepackage{makecell}
\usepackage[dvipsnames]{xcolor}
\usepackage{adjustbox} 

\renewcommand{\thefootnote}
{\fnsymbol{footnote}}

\title{Judging Against the Reference: Uncovering Knowledge-Driven Failures in LLM-Judges on QA Evaluation}

\author{Dongryeol Lee \textsuperscript{1} \hspace{1cm} Yerin Hwang\textsuperscript{2, 3} 
\hspace{1cm} Taegwan Kang \textsuperscript{2}\\
{\bf Minwoo Lee\textsuperscript{2}} 
\hspace{1.2cm} {\bf Younhyung Chae\textsuperscript{3}} 
\hspace{1.5cm}{\bf Kyomin Jung\textsuperscript{1,3$\dagger$}}\\
  \textsuperscript{1}Dept. of ECE, Seoul National University, \textsuperscript{2}LG AI Research,
  \textsuperscript{3}IPAI, Seoul National University\\
  \texttt{\{drl123, yhchae0811, kjung\}@snu.ac.kr}\\ \texttt{\{yerin.hwang, taegwan93.kang, minwoo.lee\}@lgresearch.ai}\\}

\begin{document}
\maketitle
\footnotetext{\textsuperscript{$\dagger$} Corresponding author.}

\renewcommand*{\thefootnote}
{\arabic{footnote}}
\setcounter{footnote}{0}

\begin{abstract}
While large language models (LLMs) are increasingly used as automatic judges for question answering (QA) and other reference-conditioned evaluation tasks,  little is known about their ability to adhere to a provided reference.
We identify a critical failure mode of such reference-based LLM QA evaluation: when the provided reference conflicts with the judge model's parametric knowledge, the resulting scores become unreliable, substantially degrading evaluation fidelity.
To study this phenomenon systematically, we introduce a controlled swapped-reference QA framework that induces reference–belief conflicts. Specifically, we replace the reference answer with an incorrect entity and construct diverse pairings of original and swapped references with correspondingly aligned candidate answers.
Surprisingly, grading reliability drops sharply under swapped references across a broad set of judge models.
We empirically show that this vulnerability is driven by judges' over-reliance on parametric knowledge, leading judges to disregard the given reference under conflict.
Finally, we find that this failure persists under common prompt-based mitigation strategies, highlighting a fundamental limitation of LLM-as-a-judge evaluation and motivating reference-based protocols that enforce stronger adherence to the provided reference.

\end{abstract}

\section{Introduction}
The evaluation of large language models (LLMs) is rapidly shifting toward the \emph{LLM-as-a-judge} paradigm~\cite{adlakha2024evaluating, liu2023g, gu2025surveyllmasajudge, zheng2023judging}.
Owing to their scalability and strong correlation with human annotators, LLM-as-a-judge methods have rapidly become a dominant paradigm for QA evaluation~\cite{zhang2024large, blandon2025memerag, hosseini2024benchmark, ho2025llm}, where a judge model grades a candidate answer conditioned on the question and a reference answer.
Despite these practical advantages, prior work has documented important limitations of LLM-based QA evaluation, including systematic biases and inconsistent verdicts~\cite{lee2025llm, kamalloo2024towards, li2025generation}.

\begin{figure}[t]
\centering
\includegraphics[width= 0.95\columnwidth]{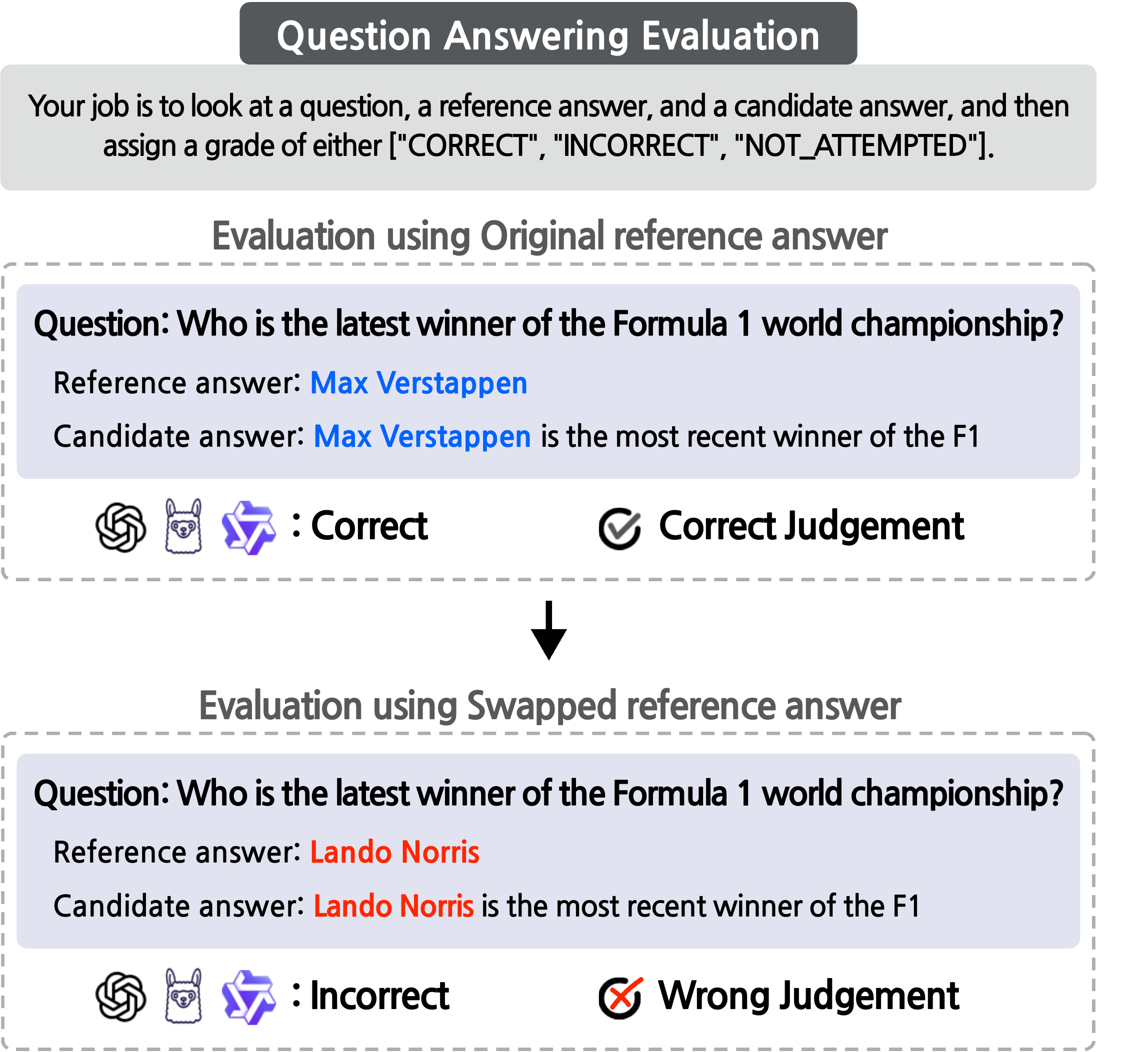} 
\caption{The LLM judge shows a critical failure mode under a swapped reference answer: with the original reference (top), it produces the correct judgment, whereas with the swapped reference (bottom), it produces an incorrect judgment even in a trivial case.}
\label{figure1}
\vspace{-4mm}
\end{figure}

However, LLM-judges' ability to adhere to a provided reference remains poorly understood, even though such adherence is central to reference-conditioned evaluation in practice.
In this paper, we investigate a striking and consequential failure mode: when the reference answer is intentionally swapped to a different entity, LLM judges can fail even in trivial cases under the provided reference.
For example, as illustrated in Figure~\ref{figure1}, a judge labels a candidate as \texttt{Incorrect} despite the candidate clearly matching the swapped reference (e.g., ``Lando Norris'').
This behavior suggests that, under swapped references, judges do not reliably follow evaluation instructions and frequently disregard the given gold reference answer.
This concern is particularly important because factual knowledge evolves over time~\cite{vu2024freshllms, kasai2023realtime, liska2022streamingqa, zhang2023large}, and datasets may intentionally encode counterfactual targets that diverge from a model's parametric knowledge~\cite{yu2023ifqa, neeman2023disentqa, wu2024reasoning}.
Motivated by this observation, we study why such failures arise in straightforward evaluation cases and aim to clarify the mechanisms driving this breakdown.

To enable controlled investigation, we propose a systematic framework for swapped-reference QA evaluation.
Specifically, we curate four existing QA datasets---NaturalQuestions-Open~\cite{lee2019latent}, PopQA~\cite{mallen2023not}, SciQ~\cite{welbl2017crowdsourcing}, and FreshQA~\cite{vu2024freshllms}---and construct tailored evaluation instances by creating diverse combinations of original and swapped reference answers together with corresponding candidate answers.
Experiments with this framework reveal that even strong judge models frequently fail to prioritize the provided reference when it is swapped.
We further show that this vulnerability persists across datasets and across different swapped entities, and does not vanish with increased evaluator capacity.

Moreover, through extensive empirical analysis, we show that this vulnerability is driven by judges' over-reliance on parametric knowledge.
Failures are strongly associated with the judge's internal beliefs: vulnerability diminishes when the swapped reference aligns with the judge's parametric knowledge, and it increases with the popularity of the associated knowledge.
We further find that common prompt-based mitigation strategies, including chain-of-thought prompting, in-context learning, and detailed task instructions, do not resolve the issue.
Taken together, these results expose a critical challenge for LLM-as-a-judge approaches in QA evaluation: when the provided gold reference conflicts with a judge’s parametric knowledge, adherence to the reference can substantially degrade.

In summary, we examine the challenges LLM judges face when the provided reference contradicts their internal knowledge, and we make the following contributions:
\begin{itemize}
    \item We uncover a critical failure mode in LLM-judge--based QA evaluation: judges often ignore evaluation instructions when the reference answer is different from models' parametric knowledge.
    \item We introduce a systematic swapped-reference QA evaluation framework that enables fine-grained and controlled analysis of LLM judges’ adherence to references across datasets, swap types, and candidate-reference alignments.
    \item We identify key factors underlying the failure, showing that stronger parametric knowledge can intensify ignorance of swapped references, and that standard prompting-based mitigation techniques provide limited relief.
\end{itemize}

\begin{figure*}[t]
\centering
\includegraphics[width= 2\columnwidth]{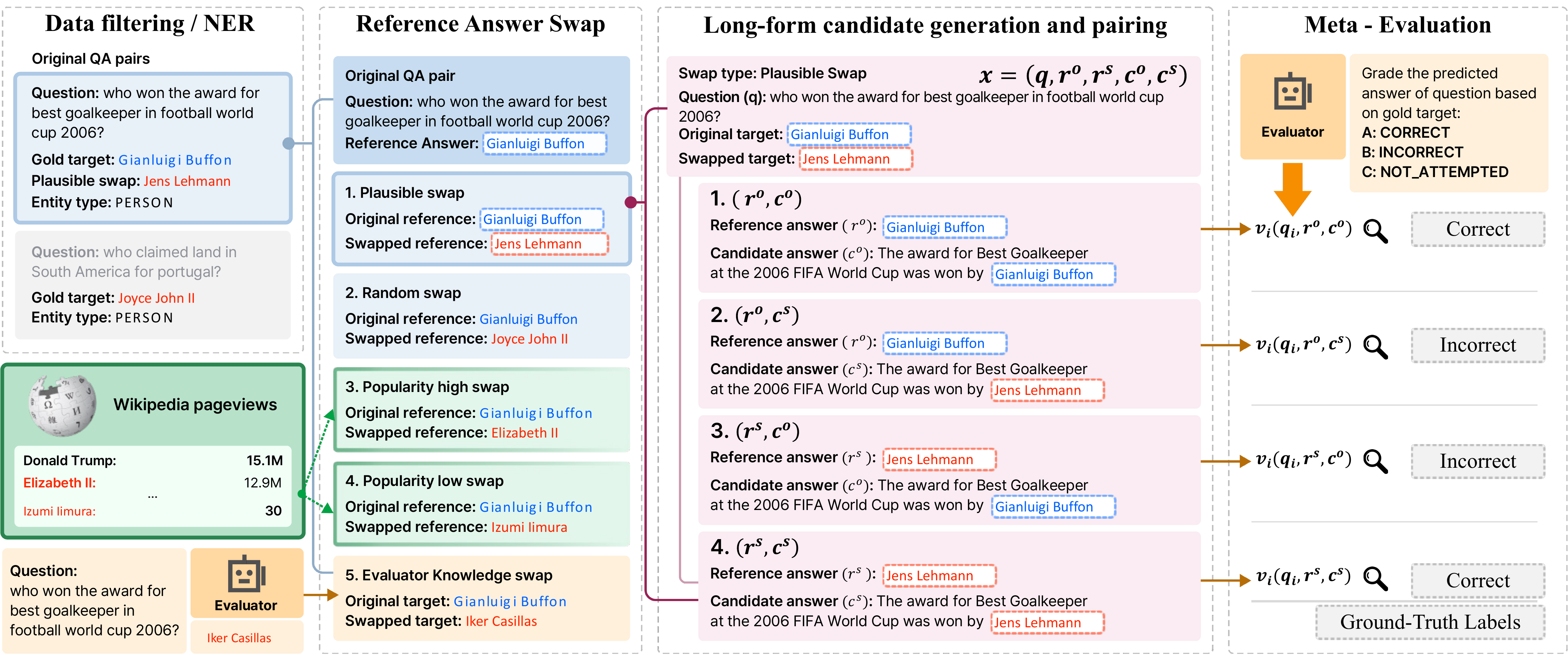} 
\caption{Evaluation framework overview}
\vspace{-3mm}
\label{figure2}
\end{figure*}

\section{Background \& Study Formulation}
\label{sec_2}

In this section, we review the standard protocol for LLM-judge--based QA evaluation and formalize the setting studied in this work.

\subsection{Background}
\label{sec_2.1}

We consider a Question Answering (QA) dataset consisting of questions ${q_i}_{i=1}^{N}$, where for each question $q_i$, the dataset provides a gold reference answer $r_i$ and the system under evaluation produces a candidate answer $c_i$.
While some QA benchmarks include multiple references per question, for simplicity we assume a single gold reference $r_i$.

Given $(q_i, r_i, c_i)$, an LLM-judge is prompted with an evaluation instruction and asked to determine whether the candidate answer is correct with respect to the \emph{provided} reference answer. The judge returns a verdict
\[
v_i = \mathrm{LLM}(q_i, r_i, c_i).
\]
These verdicts are compared against human judgments and aggregated across the dataset.
An ideal LLM-judge should follow the given instruction and condition on the provided reference, marking correct candidates as \textit{Correct} and incorrect candidates as \textit{Incorrect}.

\subsection{Swapped-Reference Setting}
\label{sec_2.2}

In standard QA benchmarks, gold reference answers are fixed at dataset creation time.
In practice, however, knowledge can evolve over time~\cite{vu2024freshllms, chen2021dataset} or vary across regions~\cite{zhang2021situatedqa}, and thus what is commonly regarded as the ``correct'' answer may shift.
Accordingly, under evaluation protocols where correctness is defined solely by agreement with the \emph{provided reference}, an ideal LLM-judge should rely on the provided reference rather than defaulting to its own prior knowledge.

To probe whether judges exhibit this reference-conditioned behavior, we introduce the \textit{Swapped-Reference} setting, in which each original answer is substituted with an alternative reference answer.
For each sample index $i$, we denote the original reference by $r_i^{o}$ and the swapped reference by $r_i^{s}$, where the superscripts $o/s$ indicate \emph{original} vs.\ \emph{swapped}.
We then examine whether the judge’s verdict changes when $r_i^{o}$ is replaced by $r_i^{s}$ thereby quantifying the extent to which LLM-judges genuinely condition on the provided reference.

\section{Experimental Setup}
\label{sec_3}

To study how LLM judges evaluate candidate answers under swapped references, this section describes our overall framework for swapped-reference QA meta-evaluation and the associated metrics.

\subsection{Swapped-Reference QA Meta-Evaluation}
\label{sec_3.1}

Each sample $x_i$ in our dataset is a quintuple
\[
x_i = (q_i, r_i^{o}, r_i^{s}, c_i^{o}, c_i^{s}),
\]
where $q_i$ is a question, $r_i^{o}$ is the original reference answer from the source dataset, $r_i^{s}$ is an artificially swapped reference answer, $c_i^{o}$ is a candidate answer aligned with $r_i^{o}$, and $c_i^{s}$ is a candidate answer aligned with $r_i^{s}$.
To reflect common QA evaluation settings, where judges often grade sentence-level or long-form responses, we generate both $c_i^{o}$ and $c_i^{s}$ as sentence-level answers.
The overall construction pipeline is depicted in Figure~\ref{figure2}; we summarize each stage below.

\paragraph{Data filtering and entity labeling (NER).}

The data filtering and NER stage in Figure~\ref{figure2} first sample questions and reference answers from original datasets.
We then apply an LLM-based named entity recognition (NER) pipeline to assign an entity-type label to each original reference answer $r_i^{o}$.

\paragraph{Reference swapping.}



Next, we perform reference swapping according to predefined strategies.
In the \emph{Random Swap} (RS) setting, we replace $r_i^{o}$ with another answer 
sampled from the same dataset that shares the same entity type (e.g., Gianluigi 
Buffon $\rightarrow$ Joyce John II in Figure~\ref{figure2}).
In the \emph{Plausible Swap} (PS) setting, we replace $r_i^{o}$ with a plausible 
but incorrect answer generated by GPT-5 (e.g., Gianluigi Buffon $\rightarrow$ 
Jens Lehmann in Figure~\ref{figure2})~\footnote{Note that GPT-5 also serves as one of the judge models evaluated in our experiments (Section~\ref{sec_3.4}). However, this does not introduce bias, as we only retain swapped references that all thirteen judges, including GPT-5, identified as incorrect}.
We also introduce a \emph{Popularity-High/Low Swap} for \textsc{person} entities.
We select the top and bottom 50 \textsc{person} entities from  PopQA~\cite{mallen2023not}, ranked by Wikipedia pageviews, and use one of them to swap \textsc{person}-type references across all datasets (e.g., Gianluigi Buffon $\rightarrow$ Elizabeth II and $\rightarrow$ Izumi Iimura, respectively, in Figure~\ref{figure2}).
For all three swap types, we retain only instances where each evaluator's predicted  answer disagrees with $r_i^{o}$, ensuring the swapped reference conflicts with the model's parametric knowledge.

Finally, we introduce an \emph{Evaluator-Knowledge Swap}.
For each evaluator model, we first query the evaluator on $q_i$ (under a standard QA prompting setup) and compare its predicted answer with $r_i^{o}$.
If the prediction disagrees with $r_i^{o}$, we use the evaluator’s predicted answer as the alternative reference and set it as the swapped reference $r_i^{s}$ for that instance.
Evaluator-Knowledge Swap explicitly aligns the swapped reference with the evaluator’s own belief.
This setting therefore isolates how judge behavior changes under knowledge alignment versus knowledge conflict.

\paragraph{Long-form candidate generation and pairing.}

Finally, we instantiate evaluation triplets by pairing each question $q_i$ with a reference choice and a candidate choice.
For each instance, we generate long-form candidates conditioned on either $r_i^{o}$ or $r_i^{s}$, yielding $c_i^{o}$ (aligned with $r_i^{o}$) and $c_i^{s}$ (aligned with $r_i^{s}$).

\paragraph{Meta-evaluation.}

We then form the four evaluation triplets, which induce ground-truth labels:
\[
\begin{aligned}
\textbf{Correct: }&(q_i, r_i^{o}, c_i^{o}),\ (q_i, r_i^{s}, c_i^{s});\\
\textbf{Incorrect: }&(q_i, r_i^{o}, c_i^{s}),\ (q_i, r_i^{s}, c_i^{o}).
\end{aligned}
\]
Thus, an ideal evaluator should consistently label aligned candidates as \textit{Correct} and misaligned candidates as \textit{Incorrect}, regardless of whether the reference is original or swapped.
We compare each evaluator’s verdict to the corresponding ground-truth label and report meta-evaluation results for each LLM judge.

Finally, human annotators manually reviewed all NER outputs, swap operations, and long-form generations to ensure that each instance is correctly swapped and that each $(q_i, r_i^{a}, c_i^{b})$ triplet carries the appropriate ground-truth label (where $a,b \in \{o,s\}$).

\begin{table*}[t]
\centering
\small
\setlength{\tabcolsep}{3pt}
\begin{tabular}{ll*{8}{l}}
\toprule
 & & \multicolumn{2}{c}{NQ-Open} & \multicolumn{2}{c}{SciQ} & \multicolumn{2}{c}{PopQA} & \multicolumn{2}{c}{FreshQA} \\
\cmidrule(lr){3-4} \cmidrule(lr){5-6} \cmidrule(lr){7-8} \cmidrule(lr){9-10}
Evaluator & Ref. &  RS & PS & RS & PS & RS & PS & RS & PS\\
\midrule
\multirow{2}{*}{Llama-3.3-70B}
 & Orig. & 97.6 & 96.3 & 96.0 & 92.0 & 99.2 & 97.1 & 96.5 & 98.8 \\
 & Swap  & 64.2 \textcolor{red}{\scriptsize(-33.4)} & 92.8 \textcolor{red}{\scriptsize(-3.5)} & 36.2 \textcolor{red}{\scriptsize(-59.8)} & 78.8 \textcolor{red}{\scriptsize(-13.2)} & 85.7 \textcolor{red}{\scriptsize(-13.5)} & 96.9 \textcolor{red}{\scriptsize(-0.2)} & 70.3 \textcolor{red}{\scriptsize(-26.2)} & 97.4 \textcolor{red}{\scriptsize(-1.4)} \\
\midrule
\multirow{2}{*}{Qwen-2.5-32B}
 & Orig. & 96.7 & 93.6 & 95.0 & 91.0 & 98.6 & 97.1 & 96.2 & 98.5 \\
 & Swap  & 66.4 \textcolor{red}{\scriptsize(-30.4)} & 90.5 \textcolor{red}{\scriptsize(-3.2)} & 45.7 \textcolor{red}{\scriptsize(-49.3)} & 74.5 \textcolor{red}{\scriptsize(-16.5)} & 87.5 \textcolor{red}{\scriptsize(-11.2)} & 95.9 \textcolor{red}{\scriptsize(-1.3)} & 72.2 \textcolor{red}{\scriptsize(-24.1)} & 98.1 \textcolor{red}{\scriptsize(-0.3)} \\
\midrule
\multirow{2}{*}{Qwen-2.5-72B}
 & Orig. & 93.9 & 91.6 & 92.7 & 84.7 & 97.3 & 95.9 & 93.4 & 95.5 \\
 & Swap  & 69.8 \textcolor{red}{\scriptsize(-24.2)} & 86.4 \textcolor{red}{\scriptsize(-5.3)} & 35.1 \textcolor{red}{\scriptsize(-57.6)} & 69.1 \textcolor{red}{\scriptsize(-15.7)} & 90.0 \textcolor{red}{\scriptsize(-7.3)} & 95.6 \textcolor{red}{\scriptsize(-0.3)} & 77.2 \textcolor{red}{\scriptsize(-16.2)} & 93.5 \textcolor{red}{\scriptsize(-2.0)} \\
\midrule
\multirow{2}{*}{Qwen-3-30B}
 & Orig. & 98.5 & 95.7 & 98.6 & 91.7 & 99.5 & 97.9 & 98.1 & 98.9 \\
 & Swap  & 69.7 \textcolor{red}{\scriptsize(-28.9)} & 89.9 \textcolor{red}{\scriptsize(-5.8)} & 49.3 \textcolor{red}{\scriptsize(-49.3)} & 72.6 \textcolor{red}{\scriptsize(-19.1)} & 86.5 \textcolor{red}{\scriptsize(-13.0)} & 96.5 \textcolor{red}{\scriptsize(-1.4)} & 69.7 \textcolor{red}{\scriptsize(-28.4)} & 97.2 \textcolor{red}{\scriptsize(-1.7)} \\
\midrule
\multirow{2}{*}{Qwen-3-30B-think}
 & Orig. & 98.7 & 98.7 & 99.0 & 99.2 & 99.7 & 99.4 & 97.7 & 98.8 \\
 & Swap  & 91.5 \textcolor{red}{\scriptsize(-7.2)}
 & 96.6 \textcolor{red}{\scriptsize(-2.1)}
 & 85.2 \textcolor{red}{\scriptsize(-13.8)}
 & 95.7 \textcolor{red}{\scriptsize(-3.5)}
 & 97.8 \textcolor{red}{\scriptsize(-1.9)}
 & 99.5 \scriptsize(+0.1)
 & 93.9 \textcolor{red}{\scriptsize(-3.8)}
 & 98.9 \scriptsize(+0.1) \\
\midrule
\multirow{2}{*}{GPT-4o}
 & Orig. & 96.5 & 96.9 & 98.5 & 99.1 & 98.0 & 97.8 & 97.1 & 98.0 \\
 & Swap  & 59.3 \textcolor{red}{\scriptsize(-37.3)} & 83.9 \textcolor{red}{\scriptsize(-13.0)} & 49.8 \textcolor{red}{\scriptsize(-48.7)} & 77.1 \textcolor{red}{\scriptsize(-22.0)} & 70.9 \textcolor{red}{\scriptsize(-27.1)} & 87.0 \textcolor{red}{\scriptsize(-10.9)} & 64.3 \textcolor{red}{\scriptsize(-32.8)} & 89.6 \textcolor{red}{\scriptsize(-8.4)} \\
\midrule
\multirow{2}{*}{GPT-5}
 & Orig. & 98.4 & 99.5 & 97.6 & 99.4 & 99.7 & 99.7 & 97.6 & 100.0 \\
 & Swap  & 95.7 \textcolor{red}{\scriptsize(-2.8)} & 98.3 \textcolor{red}{\scriptsize(-1.2)} & 95.3 \textcolor{red}{\scriptsize(-2.3)} & 98.1 \textcolor{red}{\scriptsize(-1.3)} & 99.3 \textcolor{red}{\scriptsize(-0.4)} & 99.7 \textcolor{red}{\scriptsize(0.0)} & 95.1 \textcolor{red}{\scriptsize(-2.4)} & 99.7 \textcolor{red}{\scriptsize(-0.3)} \\
\bottomrule
\end{tabular}
\caption{Accuracy (\%) of different evaluator models under Random Swap (RS) and Plausible Swap (PS) settings across datasets. Numbers in parentheses indicate the accuracy change from the original-reference setting to the swapped-reference setting; \textcolor{red}{red} denotes a decrease in accuracy.}
\vspace{-2mm}
\label{table:main}
\end{table*}

\subsection{Statistics of Datasets}
\label{sec_3.2}

For each evaluator--dataset pair, we construct a corresponding Swapped-Reference QA Meta-Evaluation subset.
Table~\ref{tab:dataset_stat} reports summary statistics for all constructed subsets across evaluators and datasets.
The number of popularity-based swaps varies by dataset, reflecting differences in the prevalence of \textsc{person} entities.
The number of Evaluator-Knowledge swaps is determined by the instances in which the evaluator’s own answer disagrees with the original reference set ${r_i^{o}}$, and therefore varies across evaluators.

\subsection{Evaluation Metric}
\label{sec_3.3}

We introduce metrics to quantify evaluator behavior under the swapped-reference setting.

\paragraph{Accuracy.}
Let $\hat{y}_i^{a,b}\in\{\textit{Correct},\textit{Incorrect}\}$ denote the evaluator's verdict for
the triplet $(q_i, r_i^{a}, c_i^{b})$, where $a,b\in\{o,s\}$.
We define the ground-truth label
\[
y_i^{a,b} =
\begin{cases}
\textit{Correct} & \text{if } a=b,\\
\textit{Incorrect} & \text{if } a\neq b.
\end{cases}
\]
Accuracy under reference condition $a\in\{o,s\}$ is defined as follows; we report $\mathrm{ACC}^{o}$ (original reference) and $\mathrm{ACC}^{s}$ (swapped reference):
\[
\mathrm{ACC}^{a}
=\frac{1}{2N}\sum_{i=1}^{N}\sum_{b\in\{o,s\}}
\mathbb{I}\!\left[\hat{y}_i^{a,b}=y_i^{a,b}\right].
\]

\paragraph{Reference-Polarity Accuracy Gap (RPAG).}

To quantify sensitivity to reference swaps, we define
\[
\mathrm{RPAG} = \big[\mathrm{ACC}^{o} - \mathrm{ACC}^{s}\big],
\]
where $\mathrm{ACC}^{o}$ and $\mathrm{ACC}^{s}$ denote accuracy under the original references ${r_i^{o}}$ and swapped references ${r_i^{s}}$, respectively.
A larger RPAG indicates that an evaluator performs substantially better when conditioned on $r_i^{o}$ but degrades under $r_i^{s}$, reflecting vulnerability under knowledge conflict.
Conversely, a near-zero RPAG suggests robustness to reference swaps and closer adherence to the provided reference.

\subsection{LLM-Evaluators}
\label{sec_3.4}
Throughout the experiments, we test thirteen LLM Judges: three GPT-family models (GPT-4o~\cite{gpt4o}, GPT-4.1~\cite{gpt4.1}, and GPT-5~\cite{gpt5}), three Llama-family models~\cite{dubey2024llama} (Llama-3.1-8B/70B, and Llama-3.3-70B), and seven Qwen-family models (Qwen-2.5-7B/32B/72B~\cite{hui2024qwen2}, Qwen-3-4B/30B/4B-think/30B-think~\cite{yang2025qwen3}).

More details of the experimental setup are provided in Appendix~\ref{app:dataset} and Appendix~\ref{app:experiment}.

\section{Are LLM-Judges Robust to Knowledge Conflicts Induced by Swapped Reference?}
\label{sec_4}

We begin with a controlled evaluation in which each LLM judge first grades candidate answers ($c_i^o, c_i^s$)  under the original reference set, yielding accuracy $ACC^{o}$, and then re-grades the same candidates under a swapped reference set, yielding accuracy $ACC^{s}$.
This design directly tests whether LLM-judge behavior changes when the provided reference is swapped.
We evaluate thirteen LLM judges across four QA datasets, considering both \textit{Random Swap} and \textit{Plausible Swap} settings (Section~\ref{sec_3.1}).

\subsection{LLM Judges Are Vulnerable to Knowledge Conflicts Induced by Swapped References}
\label{sec_4.1}
As shown in Table~\ref{table:main}, all judges exhibit substantial accuracy drops under swapped references, revealing a consistent vulnerability across models, datasets, and swap types.
Notably, even strong models such as GPT-4o and Qwen-3-30B remain vulnerable across datasets and swap configurations.
Overall, these results suggest that current LLM judges do not reliably condition their verdicts on the provided reference answer when that reference is swapped.

To further characterize this failure mode, we break down accuracy by each reference–candidate pairing in the triplet $(q_i, r_i^{a}, c_i^{b})$, with $a,b\in\{o,s\}$ in Table~\ref{tab:all}.
Accuracy drop is particularly severe for $(q_i, r_i^{s}, c_i^{s})$, when both the reference and the candidate answer are swapped.
This pattern indicates that LLM judges often reject a candidate answer even when it matches the provided reference.

\begin{figure*}[t]
\centering
\includegraphics[width=0.90\textwidth]{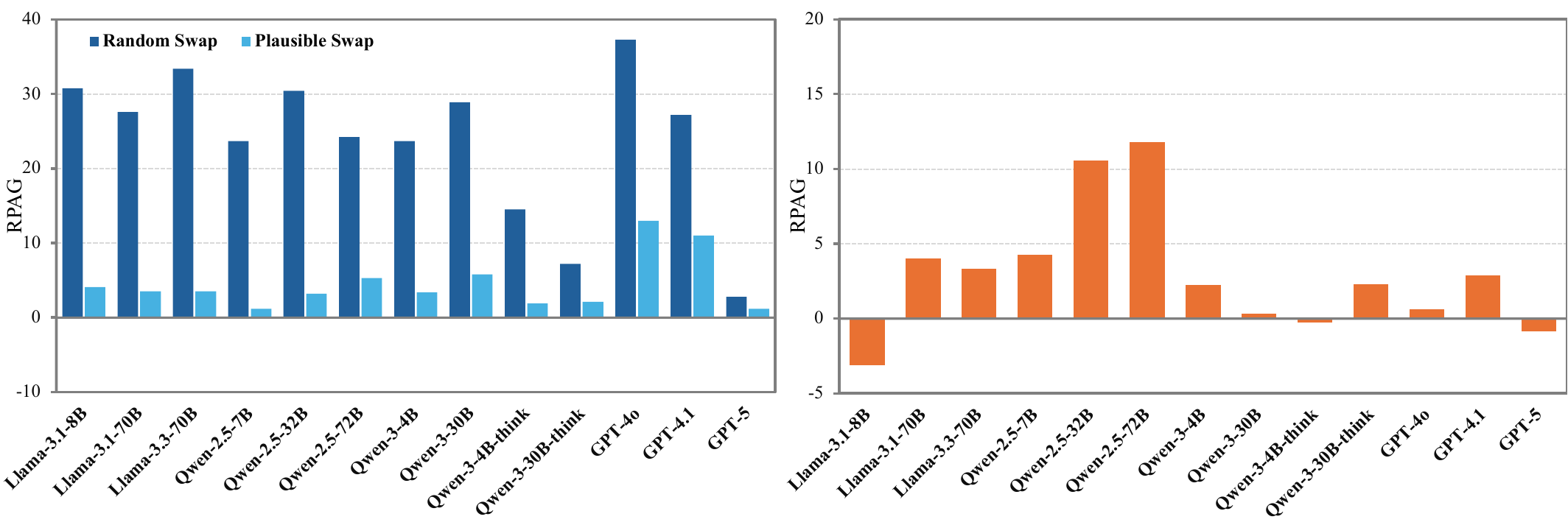}
\caption{Left: RPAG under Random Swap and Plausible Swap settings. Right: RPAG under the Evaluator-Knowledge swap setting. Results are on NQ-Open. Near-zero RPAG indicates better robustness.}
\label{figure3}
\vspace{-4mm}
\end{figure*}

\subsection{The Degree of Vulnerability Varies Across Swap Types and Datasets}
\label{sec_4.2}

Next, we examine how vulnerability varies across swap types and datasets.
The left panel of Figure~\ref{figure3} shows that RPAG is generally larger under \textit{Random Swap} than under \textit{Plausible Swap}, indicating that the accuracy drop from the original-reference setting to the swapped-reference setting is more pronounced when the swap induces higher knowledge conflict.
One plausible explanation is that \textit{Random Swap} tends to produce a higher degree of knowledge conflict (e.g., a question about the best goalkeeper may be paired with Joyce John II, who is neither a goalkeeper nor a football player (Figure~\ref{figure2})), which may trigger stronger model priors and increase the likelihood that the judge disregards the provided reference.

As shown in Table~\ref{table:main}, the degree of accuracy degradation differs considerably across datasets.
In particular, PopQA exhibits consistently smaller accuracy drops, whereas SciQ shows markedly larger degradations for most models.
This discrepancy can be attributed to differences in question characteristics.
PopQA largely consists of ``who'' questions, which tend to preserve semantic coherence even after reference swapping, thereby partially masking the underlying vulnerability.
In contrast, SciQ primarily probes precise scientific knowledge that is strongly encoded in evaluators' parametric knowledge and remains stable over time, increasing the likelihood that evaluators disregard the provided reference.


\subsection{Vulnerability Persists with Increased Model Capacity}
\label{sec_4.3}

Finally, we test whether scaling model capacity improves robustness to swapped references.
Intuitively, larger models might better follow instructions or more effectively separate reference-conditioned evaluation from prior knowledge.

However, our results show that increasing model size does not mitigate this vulnerability.
As shown in Figure~\ref{figure3}, across a range of parameter scales, RPAG does not systematically diminish with scale.
In some cases, larger models even exhibit greater vulnerability (e.g., Qwen-2.5-32B and Qwen-3-30B show higher RPAG than Qwen-2.5-7B and Qwen-3-4B, respectively), suggesting that increased capacity may strengthen reliance on internal world knowledge rather than improve adherence to the provided reference.
Taken together, these findings indicate that reference-disregarding behavior is not merely a consequence of limited model capacity, but instead reflects a more fundamental limitation of current LLM judges.

\section{Why Do LLM-Judges Ignore Swapped Reference Answers?}

In this section, we investigate the mechanisms underlying the vulnerability of LLM judges observed in Section~\ref{sec_4}.
We first test whether this failure is associated with judges relying on parametric knowledge rather than adhering to the provided reference (\S~\ref{sec_5.1}).
We then examine whether the popularity of the  knowledge associated the swapped reference correlates with the severity of the vulnerability (\S~\ref{sec_5.2}).
Finally, we analyze how question characteristics relate to vulnerability, with a particular focus on knowledge freshness of question (\S~\ref{sec_5.3}).

\subsection{Vulnerability Diminishes When the Swapped Reference Aligns with Parametric Knowledge}
\label{sec_5.1}

We begin by testing whether the observed vulnerability reflects a systematic preference for parametric knowledge over the provided reference.
Under the \textit{Evaluator-Knowledge Swap} setting (Section~\ref{sec_3.1}), we replace the reference answer with an alternative that matches the judge's internal beliefs in instances where those beliefs conflict with the original reference.
As shown in Figure~\ref{figure3}, when the reference is swapped to align with the evaluator's parametric knowledge, RPAG becomes nearly zero: swapped-reference accuracy ($ACC^{s}$) becomes nearly identical to original-reference accuracy ($ACC^{o}$).

This pattern suggests that LLM judges can execute the evaluation procedure reliably when the reference does not contradict their internal knowledge.
Therefore, these failures are less likely due to instruction- or format-following limitations, and instead reflect a conflict-resolution tendency where the judge prioritizes its parametric knowledge when the provided reference conflicts with it.

\subsection{Swapped-Reference Popularity Correlates with Vulnerability}
\label{sec_5.2}

Next, we analyze how the popularity of knowledge associated with the swapped reference entity affects vulnerability.
Following Section~\ref{sec_3.1}, we evaluate Popularity-high versus Popularity-low swaps on NQ-Open and PopQA.\footnote{Popularity swaps are only applicable to \textsc{person} entities. We therefore report results on NQ-Open and PopQA, which contain sufficient numbers of \textsc{person} instances (359 and 585, respectively).}

\begin{figure}[t]
\centering
\includegraphics[width= 0.90\columnwidth]{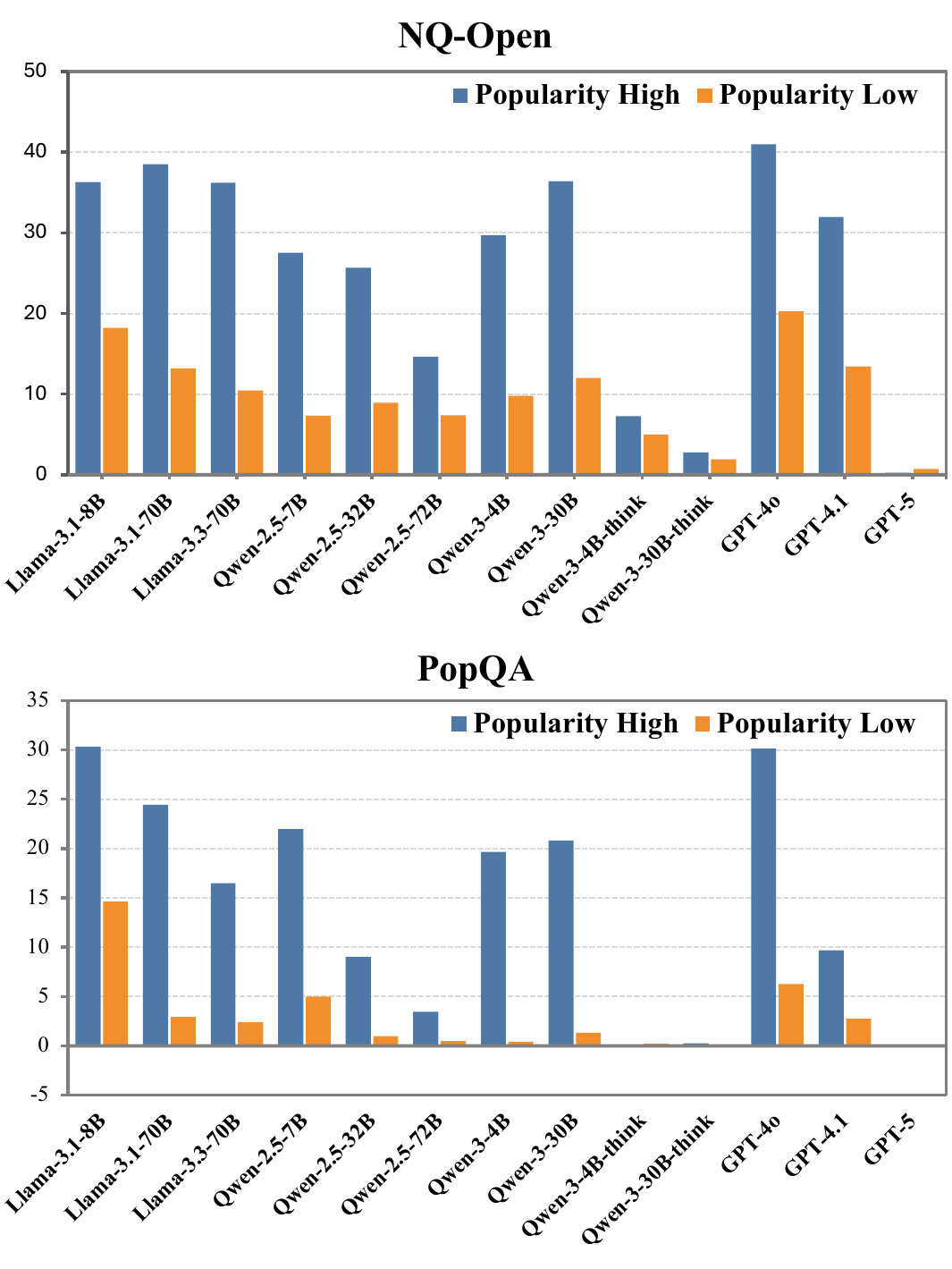} 
\caption{RPAG under Popularity-High and Popularity-Low swap settings on the NQ-Open and PopQA datasets. Near-zero RPAG indicates better robustness.}
\label{figure4}
\vspace{-4mm}
\end{figure}

As shown in Figure~\ref{figure4}, vulnerability is substantially more severe under Popularity-high swaps than under Popularity-low swaps across all evaluators.
A plausible explanation is that Popularity-high swaps introduce widely known entities for which the evaluator has strong associations, increasing the likelihood of a direct reference--belief conflict.
In such cases, the judge appears more likely to discount the provided reference and revert to its parametric knowledge.
Conversely, when the reference is swapped to a less popular entity, the evaluator may have weaker or more uncertain priors, reducing the probability of a conflict and thereby attenuating the vulnerability.
These observations align with \S~\ref{sec_5.1} and suggest that the failure severity increases with the level of parametric knowledge tied to the swapped answer.

\subsection{Knowledge Freshness Correlates with Vulnerability}
\label{sec_5.3}

Finally, we examine whether vulnerability varies with knowledge freshness.
FreshQA~\cite{vu2024freshllms} provides question-level labels based on how likely the underlying fact is to change over time: \textit{Never-changing}, \textit{Slow-changing}, and \textit{Fast-changing}.
Intuitively, faster-changing facts are less likely to be strongly encoded in an LLM judge's parametric knowledge.

\begin{figure}[t]
\centering
\includegraphics[width= 0.90\columnwidth]{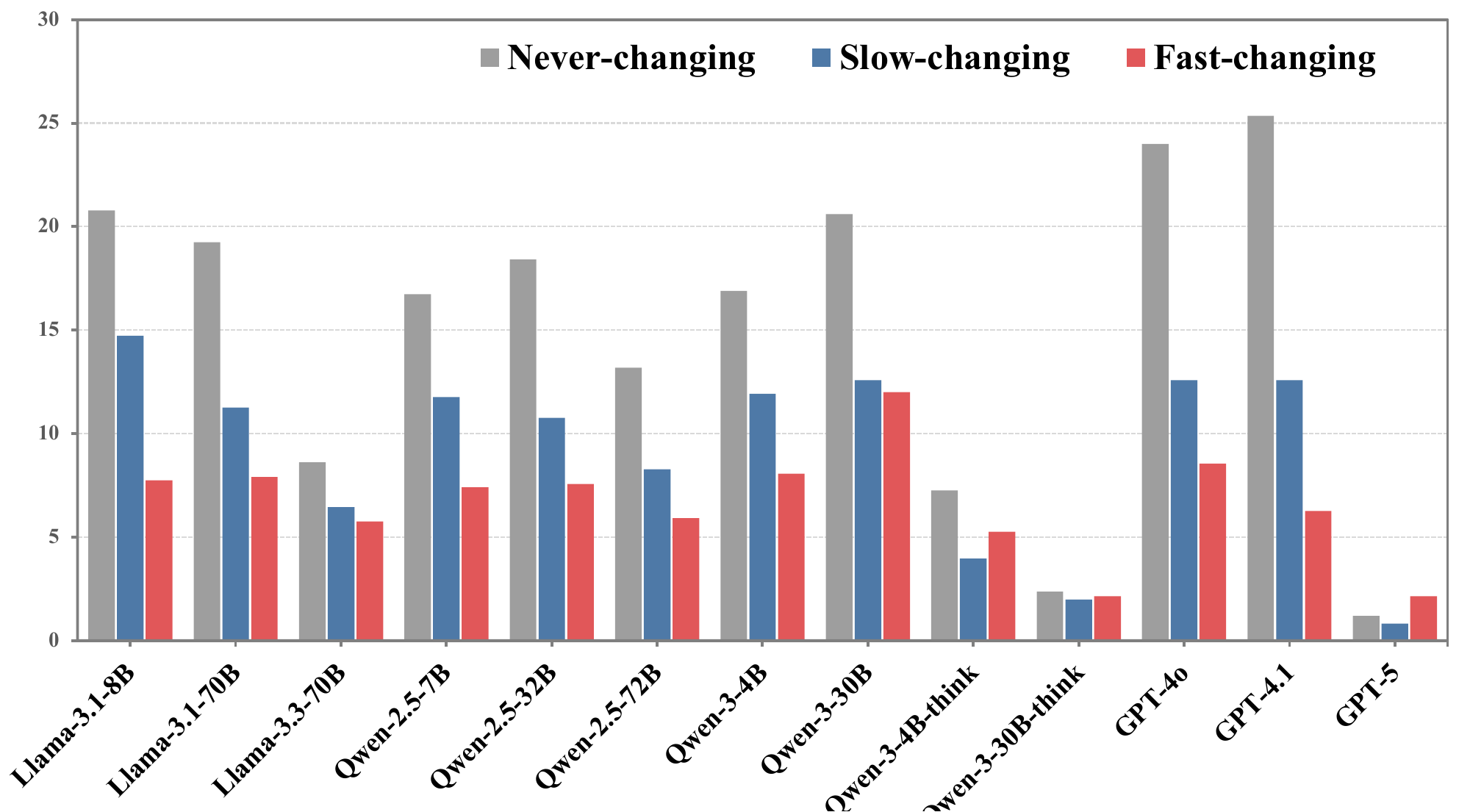} 
\caption{RPAG under Random Swap and Plausible Swap settings across question freshness types (never-, slow-, and fast-changing), evaluated on FreshQA. Near-zero RPAG indicates better robustness.}
\label{figure_freshness}
\vspace{-4mm}
\end{figure}

Using the freshness labels provided by FreshQA, we stratify the average RPAG scores for Random Swap and Plausible Swap by freshness level.
As shown in Figure~\ref{figure_freshness}, questions involving rapidly changing or time-sensitive knowledge show low RPAG, indicating lower vulnerability under swapped references.
One interpretation is that, for such questions, LLM judges have weaker or lower-confidence parametric knowledge, making them more willing to defer to the provided reference.
In contrast, questions about stable, long-standing facts tend to exhibit higher vulnerability, consistent with stronger parametric beliefs that are more resistant to being overridden by the reference.

Taken together, these results indicate that susceptibility to swapped references is shaped by the interaction between the level of the judge’s parametric knowledge and the extent to which the provided reference is treated as reliable under knowledge conflict.

\section{Can Prompting Mitigate the Vulnerability?}
\label{sec_6}
Prior works suggest that prompting strategies~\cite{zhu2023judgelm, hwang2025can, chen2024factors, kamalloo2024towards}, can improve evaluator robustness.
We examine whether these prompting strategies mitigate LLM-judge vulnerability under knowledge conflict by evaluating four approaches: \textit{Direct}, \textit{In-context Learning (ICL)}, \textit{CoT}~\cite{wei2022chain}, and \textit{CoT+Self-consistency}.
All experiments use GPT-4o under both Random Swap and Plausible Swap settings across four datasets, with prompt details provided in Appendix~\ref{app:experiment}.

As shown in Table~\ref{tab:prompting}, none of the tested strategies fully eliminates the vulnerability.
Among them, \textit{Direct} reduces (but does not remove) the vulnerability relative to the standard prompting baseline, suggesting that stronger instructions can partially improve reference adherence but remains insufficient under conflict.
Notably, \textit{CoT} exhibits asymmetric behavior across conflict conditions: it reduces vulnerability under \textit{Plausible Swap} but increases vulnerability under \textit{Random Swap}. 
As analyzed in Appendix~\ref{app:reasoning}, eliciting explicit reasoning tends to amplify reliance on the judge's parametric beliefs when the degree of knowledge conflict is large (\textit{Random Swap}), increasing the likelihood that the model overrides the provided reference. 
This behavior contrasts with prior findings that CoT, particularly with self-consistency, consistently improves reliability in standard QA evaluation settings~\cite{kamalloo2024towards}.
Overall, these results indicate that prompting strategies effective under non-conflicting references settings do not generalize to knowledge-conflict settings.
\begin{table}[t]
\centering
\small
\setlength{\tabcolsep}{2.5pt}
\resizebox{\columnwidth}{!}{%
\begin{tabular}{l*{8}{r}}
\toprule
 & \multicolumn{2}{c}{NQ-Open} & \multicolumn{2}{c}{SciQ} & \multicolumn{2}{c}{PopQA} & \multicolumn{2}{c}{FreshQA} \\
\cmidrule(lr){2-3}\cmidrule(lr){4-5}\cmidrule(lr){6-7}\cmidrule(lr){8-9}
Prompting & RS & PS & RS & PS & RS & PS & RS & PS \\
\midrule
Standard         & 37.3 & 13.0 & 48.7 & 22.0 & 27.1 & 10.9 & 32.8 &  8.4 \\
Direct           & 22.5 & 11.2 & 34.1 & 20.2 & 18.5 & 10.4 & 21.5 &  7.4 \\
ICL              & 33.3 & 15.6 & 49.2 & 26.6 & 25.9 & 12.4 & 28.4 &  9.2 \\
CoT              & 40.3 &  8.1 & 66.2 & 17.8 & 15.7 &  2.3 & 35.0 &  2.1 \\
CoT+Self-Consist & 42.7 &  7.1 & 65.4 & 11.8 & 16.7 &  2.0 & 39.6 &  1.9 \\
\bottomrule
\end{tabular}%
}
\caption{RPAG under Random Swap (RS) and Plausible Swap (PS) settings across datasets for different prompting strategies using GPT-4o.}
\vspace{-1mm}
\label{tab:prompting}
\end{table}


\section{Related Works}
\label{sec_7}
\subsection{LLM-as-a-Judge}
LLM-as-a-judge has emerged as a scalable paradigm for evaluating model outputs and often correlates well with human judgments~\cite{zheng2023judging, wang2023chatgpt, liu2023g, chiang2023closer, gu2025surveyllmasajudge, thakur2025judging}. 
In QA evaluation, traditional metrics such as EM, F1, and BLEU~\cite{papineni2002bleu} are increasingly misaligned with long-form, semantically diverse answers produced by modern LLMs~\cite{kamalloo2023evaluating, wang2023evaluating, lee2025return}, motivating broad adoption of LLM-judges for reference-conditioned evaluation~\cite{zhang2024large, blandon2025memerag, hosseini2024benchmark, ho2025llm, chen2024factors}.

At the same time, prior work documents systematic limitations of LLM-based evaluation~\cite{li2025generation}, including biases toward longer outputs~\cite{koo2024benchmarking, dubois2024length}, well-formatted responses~\cite{chen2024humans, stephan2025calculation}, and answers without epistemic markers~\cite{lee2025llm}. 
Another recurring concern is self- or model-family preference, where judges favor their own generations or closely related models~\cite{liu2024llms, chen2025beyond}. 
These findings have motivated diverse meta-evaluations of LLM-judges that systematically probe when they deviate from their intended grading criteria~\cite{tan2024judgebench, zheng2023judging}.

\subsection{Knowledge Conflict in LLMs}
Knowledge conflict arises because LLMs increasingly combine parametric knowledge stored in model weights with contextual knowledge supplied at inference time~\cite{pan2022knowledge}, including user instructions~\cite{liu2023pre}, retrieved documents~\cite{kim2026trainingdatashapesuse, shi2024replug}, and tool outputs~\cite{schick2023toolformer, xu2024knowledge}. 
When these sources disagree, models must implicitly resolve a conflict, and the chosen resolution strategy can substantially affect reliability. 
This issue is widely studied in QA, especially in retrieval-augmented generation (RAG), where models may under- or over-incorporate retrieved evidence~\cite{xie2023adaptive, jin2024tug, tan2024blinded}.

Several works intentionally introduce contradictions to characterize such behavior: \citet{longpre2021entity} and \citet{chen2022rich} show that models differ in their tendency to rely on parametric versus contextual information, while IfQA~\cite{yu2023ifqa} and DisentQA~\cite{neeman2023disentqa} use counterfactual context augmentation to test whether models incorporate edited evidence. 
Related analyses extend to multimodal settings by inducing conflicts between images and parametric knowledge~\cite{bitton2023breaking, zhou2023rome, guan2024hallusionbench, luo2024probing, lee2025vlind}. 
To the best of our knowledge, however, prior work has not systematically studied \emph{LLM-judges} under knowledge conflict in \emph{QA evaluation}, where judges must reconcile their own beliefs with an explicitly provided reference when grading candidate answers.

\section{Conclusion}
We show that LLM judges often fail to adhere to reference-conditioned QA evaluation when the provided reference conflicts with their parametric knowledge.
Using our swapped-reference QA evaluation framework across four datasets and multiple judge models, we find that even strong evaluators frequently fail on otherwise trivial cases, and this vulnerability persists across swap types, swap targets, and model scale.
Our analyses indicate that this vulnerability arises from judges’ over-reliance on internal beliefs, and that common prompting strategies provide only limited mitigation.
These findings highlight a fundamental risk in LLM-as-a-judge evaluation under knowledge conflict and motivate the development of protocols that enforce stronger reference adherence.

\section*{Limitations}
This work focuses exclusively on question answering (QA) evaluation.
While QA is one of the most common and impactful applications of the LLM-as-a-judge paradigm, our findings do not directly establish that the same reference-adherence failures arise in other reference-conditioned evaluation settings, such as summarization, fact verification, or dialog evaluation.
Extending the swapped-reference analysis to these broader tasks remains an important direction for future work.

Second, our study does not propose a concrete method to mitigate the observed vulnerability. Instead, our goal is diagnostic: to isolate, characterize, and analyze a fundamental failure mode of LLM judges under reference and knowledge conflict. Designing effective mitigation strategies that reliably enforce reference adherence is a non-trivial challenge and lies beyond the scope of this work. One promising direction is fine-tuning LLM judges on datasets that explicitly instantiate knowledge conflicts, such as training instances in which the correct evaluation outcome requires overriding the model's parametric beliefs in favor of the provided reference~\cite{zhu2025judgelm}. Such targeted supervision could encourage judges to treat the reference as the authoritative signal, rather than a cue to be weighed against prior knowledge.

Finally, our evaluation relies on artificially swapped reference answers, which can produce question–reference pairs that appear counterintuitive or implausible from a real-world perspective.
However, this design choice is intentional and necessary to create controlled conflicts between the provided reference and the judge’s parametric knowledge.
Crucially, the intended role of an LLM judge in reference-based QA evaluation is to follow explicit evaluation instructions and assess candidate answers with respect to the provided reference, regardless of its alignment with prior beliefs.
Therefore, failures under swapped references highlight a limitation of current LLM judges, rather than an artifact of unrealistic evaluation settings.

\section*{Ethical Considerations}
In our experiments, we utilize publicly available datasets, including NQ-Open~\cite{lee2019latent}, SciQ~\cite{welbl2017crowdsourcing}, PopQA~\cite{mallen2023not}, and FreshQA~\cite{vu2024freshllms}.
These datasets are widely adopted and well-established within the research community, and their use raises no additional privacy or consent concerns beyond those addressed in the original dataset releases.

All large language models used in this work were accessed through their official and publicly available sources.
Specifically, GPT-family models were accessed via OpenAI’s official platform\footnote{\url{https://openai.com}}, while LLaMA-family and Qwen-family models were obtained from their respective official releases under the corresponding usage licenses.
Our use of these models complies with their stated terms of service and aligns with open science and reproducibility principles.

During the preparation of this manuscript, we employed an AI-assisted writing tool at the sentence level to support drafting and linguistic refinement.
The AI assistant did not generate experimental results, design methodologies, or draw scientific conclusions, and all technical content and interpretations remain the responsibility of the authors.



\bibliography{custom}

\appendix
\newpage

\clearpage

\section{Details of Dataset}
\label{app:dataset}

We sampled datasets from NQ-Open~\cite{lee2019latent}, PopQA~\cite{mallen2023not}, Sciq~\cite{welbl2017crowdsourcing}, and FreshQA~\cite{vu2024freshllms}.
For NQ-Open, PopQA, and SciQ, we randomly sample 1{,}000 instances from the original dataset.\footnote{For NQ-Open, we randomly sample 1{,}000 instances from the validation set; for PopQA, we randomly sample 1{,}000 instances from the test set; for SciQ, we use the 1{,}000-question validation set.}
For FreshQA, we use the dev and test splits of the \textit{August 18, 2025} version and exclude questions with false premises, yielding 452 instances.

Throughout dataset construction (e.g., NER, swapping, and candidate generation), we used GPT-4o (\texttt{gpt-4o-2024-08-06})\footnote{\url{https://platform.openai.com/docs/models/gpt-4o}}
 with temperature 0 to perform NER and to generate long-form candidates, as described in Section~\ref{sec_3.1}.
This design helps mitigate \emph{egocentric bias}, where LLM judges tend to prefer their own generations~\cite{liu2024llms, chen2025beyond}.
The prompts used for NER and for generating the original and swapped long-form candidates are shown in Figures~\ref{fig:NER_prompt}, \ref{fig:orig_long_form}, and \ref{fig:swap_long_form}, respectively.

For Plausible Swap (PS), we used GPT-5 (\texttt{gpt-5-2025-08-07})%
\footnote{\url{https://platform.openai.com/docs/models/gpt-5}}
with the prompts shown in Figure~\ref{fig:plausible_swap}.
To ensure that the swapped reference conflicts with evaluators' internal knowledge, we queried each evaluator on whether the swapped reference is the correct answer for the corresponding question, and retained only swapped references that no judge among all thirteen evaluators used in our experiments agreed with.

Table~\ref{tab:dataset_stat} reports dataset statistics.
Throughout dataset construction, the authors manually inspected entity extraction, swap operations, and generated candidates to ensure data quality and reliability.

\section{Experimental Details}
\label{app:experiment}

\subsection{Judge Models}
As described in Section~\ref{sec_3.4}, we evaluated thirteen LLM judges.
For GPT-family models, we used the official OpenAI API with the following versions: GPT-4o (\texttt{gpt-4o-2024-08-06})\footnote{\url{https://platform.openai.com/docs/models/gpt-4o}}, GPT-4.1 (\texttt{gpt-4.1-2025-04-14})\footnote{\url{https://platform.openai.com/docs/models/gpt-4.1}}, and GPT-5 (\texttt{gpt-5-2025-08-07})\footnote{\url{https://platform.openai.com/docs/models/gpt-5}}.
For Llama-family and Qwen-family models, we used the official Hugging Face repositories: Llama-3.1-8B\footnote{\url{https://huggingface.co/meta-llama/Llama-3.1-8B-Instruct}}, Llama-3.1-70B\footnote{\url{https://huggingface.co/meta-llama/Llama-3.1-70B-Instruct}}, and Llama-3.3-70B\footnote{\url{https://huggingface.co/meta-llama/Llama-3.3-70B-Instruct}}; Qwen-2.5-7B\footnote{\url{https://huggingface.co/Qwen/Qwen2.5-7B-Instruct}}, Qwen-2.5-32B\footnote{\url{https://huggingface.co/Qwen/Qwen2.5-32B-Instruct}}, and Qwen-2.5-72B\footnote{\url{https://huggingface.co/Qwen/Qwen2.5-72B-Instruct}}; and Qwen-3-4B\footnote{\url{https://huggingface.co/Qwen/Qwen3-4B-Instruct-2507}}, Qwen-3-30B\footnote{\url{https://huggingface.co/Qwen/Qwen3-30B-A3B-Instruct-2507}}, Qwen-3-4B-Think\footnote{\url{https://huggingface.co/Qwen/Qwen3-4B-Thinking-2507}}, and Qwen-3-30B-Think\footnote{\url{https://huggingface.co/Qwen/Qwen3-30B-A3B-Thinking-2507}}.

\subsection{Hyperparameters}
Across all experiments, we set the decoding temperature to $0$, following common practice in LLM-as-a-judge evaluation settings~\cite{wang2024large, wang2023evaluating, lee2025llm}.
Unless otherwise specified, we used greedy decoding with fixed max generation length and the same prompt template across judges.

\subsection{Prompts}
For QA evaluation, we adopt the SimpleQA evaluation template introduced by OpenAI,%
\footnote{\url{https://openai.com/index/introducing-simpleqa/}}
as shown in Table~\ref{fig:simpleeval_prompt}.
The SimpleQA evaluation template has become the de facto standard evaluation prompt for QA in recent years.
To test whether our findings generalize across different evaluation prompts, we conduct additional experiments using two widely adopted QA evaluation prompts from prior work~\cite{kamalloo2023evaluating, wang2023evaluating}, with GPT-4o as the judge.
As shown in Table~\ref{tab:additional_prompt_result}, the vulnerability persists across different prompt templates, indicating that LLM-judge susceptibility to swapped references is not an artifact of the evaluation prompt but rather a more fundamental limitation.
\begin{table}[t]
\centering
\small
\setlength{\tabcolsep}{2.5pt}
\resizebox{\columnwidth}{!}{%
\begin{tabular}{l*{8}{r}}
\toprule
 & \multicolumn{2}{c}{NQ-Open} & \multicolumn{2}{c}{SciQ} & \multicolumn{2}{c}{PopQA} & \multicolumn{2}{c}{FreshQA} \\
\cmidrule(lr){2-3}\cmidrule(lr){4-5}\cmidrule(lr){6-7}\cmidrule(lr){8-9}
Prompting & RS & PS & RS & PS & RS & PS & RS & PS \\
\midrule
Standard         & 37.3 & 13.0 & 48.7 & 22.0 & 27.1 & 10.9 & 32.8 &  8.4 \\
Direct           & 22.5 & 11.2 & 34.1 & 20.2 & 18.5 & 10.4 & 21.5 &  7.4 \\
ICL              & 33.3 & 15.6 & 49.2 & 26.6 & 25.9 & 12.4 & 28.4 &  9.2 \\
CoT              & 40.3 &  8.1 & 66.2 & 17.8 & 15.7 &  2.3 & 35.0 &  2.1 \\
CoT+Self-Consist & 42.7 &  7.1 & 65.4 & 11.8 & 16.7 &  2.0 & 39.6 &  1.9 \\
\midrule
\citet{wang2023evaluating}      & 36.2 & 16.0 & 54.9 & 26.5 & 25.6 & 13.7 & 32.2 & 12.2 \\
\citet{kamalloo2023evaluating}  & 54.6 & 22.9 & 74.5 & 44.6 & 42.9 & 21.1 & 45.0 & 17.9 \\
\bottomrule
\end{tabular}%
}
\caption{RPAG under Random Swap (RS) and Plausible Swap (PS) settings across datasets for different prompting strategies using GPT-4o.}
\vspace{-1mm}
\label{tab:additional_prompt_result}
\end{table}

For the experiments in Section~\ref{sec_6}, we evaluate the following four prompting strategies: 
(1) \textit{Direct}, which explicitly instructs the judge to base its verdict solely on the provided reference; 
(2) \textit{CoT}, which requires the judge to produce a reasoning trace before outputting a final verdict~\cite{wei2022chain}; 
(3) \textit{CoT+Self-consistency}, which applies majority voting over CoT-based samples; and
(4) \textit{In-Context Learning}, which applies few-shot examples that reflects reference swap settings.

The prompts for \textit{Direct}, \textit{CoT}, and \textit{In-Context Learning} are shown in Tables~\ref{fig:direct_prompt}, ~\ref{fig:cot_prompt}, and ~\ref{fig:icl_prompt}, respectively.

\subsection{Computing Resources}
For the experiments, we utilize two 8 NVIDIA A100 Tensor Core GPUs (each with 80GB of memory).

\section{Additional Analysis}
\subsection{Manual Inspection of Reasoning Paths}
\label{app:reasoning}
Table~\ref{tab:prompting} shows that Chain-of-Thought (CoT) prompting exhibits asymmetric behavior across conflict conditions: it reduces vulnerability under \textit{Plausible Swap} but increases vulnerability under \textit{Random Swap}.
To better understand this behavior, we manually inspected the reasoning paths of 50 examples for the Random Swap settings.
We focus on instances where the judge produced a correct verdict under standard prompting but switched to an incorrect verdict when CoT prompting was applied.

As shown in Table~\ref{tab:cot_reasoning}, all inspected reasoning paths exhibit explicit evidence of reference override driven by the model’s parametric knowledge.
Despite being provided with the reference answer, the judges frequently justify their decisions by appealing to world knowledge that contradicts the reference, ultimately producing incorrect verdicts.
These observations provide qualitative evidence supporting our quantitative findings, demonstrating that under knowledge conflict, LLM judges systematically prioritize parametric knowledge over the provided reference, even when encouraged to reason explicitly.

\subsection{Full Results Table}
We report the detailed accuracy across different reference--candidate pairing: $(r^o,r^o)$, $(r^o,r^s)$, $(r^s,r^o)$, and $(r^s,r^s)$ in Table~\ref{tab:all}.

\begin{table}[t]
\centering
\small
\setlength{\tabcolsep}{3pt}
\renewcommand{\arraystretch}{1.05}
\resizebox{\columnwidth}{!}{%
\begin{tabular}{llrrrr}
\toprule
\multicolumn{2}{l}{\textbf{Swap Type}} & \textbf{NQ-Open} & \textbf{SciQ} & \textbf{PopQA} & \textbf{FreshQA} \\
\midrule

\textbf{Random Swap}      &  & 1000 & 996  & 1000 & 451 \\
\textbf{Plausible Swap}        &  & 1000 & 1000 & 1000 & 452 \\
\textbf{Popularity}  &  & 359  & 4    & 585  & 103 \\
\midrule
\multicolumn{6}{l}{\textbf{Evaluator Knowledge}} \\
Llama-3.1-8B     &  & 453 & 95  & 390 & 241 \\
Llama-3.1-70B    &  & 118 & 20  & 161 & 82  \\
Llama-3.3-70B    &  & 207 & 38  & 291 & 151 \\
Qwen-2.5-7B      &  & 169 & 62  & 149 & 108 \\
Qwen-2.5-32B     &  & 296 & 69  & 315 & 140 \\
Qwen-2.5-72B     &  & 310 & 59  & 314 & 114 \\
Qwen-3-4B        &  & 412 & 75  & 438 & 134 \\
Qwen-3-30B       &  & 336 & 52  & 373 & 134 \\
Qwen3-4B-think  &  & 411 & 97 & 346 & 211 \\
Qwen3-30B-think  &  & 443 & 110 & 409 & 185 \\
GPT-4o           &  & 327 & 96  & 355 & 200 \\
GPT-4.1          &  & 296 & 86  & 227 & 157 \\
GPT-5            &  & 351 & 100 & 204 & 115 \\

\bottomrule
\end{tabular}%
}
\caption{Dataset statistics used throughout the experiments. The number under Evaluator Knowledge indicates the number of instances in which the evaluator’s own answer disagrees with the original reference answer.}
\label{tab:dataset_stat}
\end{table}

\begin{table*}[t]
\centering
\scriptsize
\setlength{\tabcolsep}{5pt}
\renewcommand{\arraystretch}{1.1}
\resizebox{0.9\textwidth}{!}{%
\begin{tabular}{p{2.8cm} p{5.2cm} p{3.6cm} p{6.6cm}}
\toprule
\textbf{Swap Type} & \textbf{Question} & \textbf{Reference} & \textbf{Candidate Answer} \\
\midrule

Random Swap
& How many UC schools are there in the United States?
& $r^o$: 10
& $c^o$: There are 10 University of California (UC) campuses throughout the United States. \\
& &
  $r^s$: 126
& $c^s$: There are 126 UC schools located throughout the United States. \\
\midrule

Plausible Swap
& Iceland is made up of a series of
& $r^o$: volcanoes
& $c^o$: Iceland is composed of a series of volcanoes. \\
& &
  $r^s$: glaciers
& $c^s$: Iceland is made up of a series of glaciers. \\
\midrule

Popularity-High
& Who played the Elephant Man in the film?
& $r^o$: John Hurt
& $c^o$: The Elephant Man was portrayed by John Hurt. \\
& &
  $r^s$: Dwayne Johnson
& $c^s$: Dwayne Johnson played the Elephant Man in the film. \\
\midrule

Popularity-Low
& Who is the father of Ramkarpal Singh?
& $r^o$: Karpal Singh
& $c^o$: Ramkarpal Singh's father is Karpal Singh. \\
& &
  $r^s$: Jeong Ji-u
& $c^s$: The father of Ramkarpal Singh is Jeong Ji-u. \\
\midrule

Evaluator (GPT-4o) 

Knowledge
& Word that means separation of church and state
& $r^o$: separationism
& $c^o$: The term refers to separation of church and state. \\
& &
  $r^s$: secularism
& $c^s$: The word is secularism. \\
\bottomrule
\end{tabular}}
\caption{Examples of swapped-reference QA instances across different swap types.}
\label{tab:swap_examples}
\end{table*}

\begin{figure*}[!t]
    \centering
    \begin{minipage}{0.95\textwidth}
    \begin{tcolorbox}[
      title=Prompt for Named Entity Recognition,
      colframe=black!80!white,
      colback=gray!10,
      coltitle=white,
      colbacktitle=black!80!white,
      fonttitle=\bfseries,
      rounded corners,
      boxsep=3pt,
      width=\textwidth
    ]
    \small
    \vspace{5pt}
    \begin{tabular}{p{0.97\textwidth}}
    \toprule
    \textbf{Role Definition (System Prompt):}\\
    You are an information extraction model that assigns exactly ONE SpaCy NER type to the given original answer.
    Return ONLY the label token. No extra text.
    \\
    \midrule
    \textbf{User Prompt:}\\
    \#\# SpaCy NER labels \& definitions
    
    PERSON: People, including fictional.
    
    NORP: Nationalities or religious or political groups.
    
    FAC: Buildings, airports, highways, bridges, etc.
    
    ORG: Companies, agencies, institutions, etc.
    
    GPE: Countries, cities, states.
    
    LOC: Non-GPE locations, mountain ranges, bodies of water.
    
    PRODUCT: Objects, vehicles, foods, devices (not services).
    
    EVENT: Named hurricanes, battles, wars, sports events, festivals, etc.
    
    WORK\_OF\_ART: Titles of books, movies, songs, paintings, etc.
    
    LAW: Named documents made into laws.
    
    LANGUAGE: Any named language.
    
    DATE: Absolute or relative dates or periods.
    
    TIME: Times smaller than a day.
    
    PERCENT: Percentage, including “
    
    MONEY: Monetary values, including unit.
    
    QUANTITY: Measurements of size/weight/distance/volume/speed/etc.
    
    ORDINAL: “first”, “second”, “23rd”, etc.
    
    CARDINAL: Numerals that do not fall under another type.

    \#\# Rules
    
    1) Classify the answer string AS WRITTEN (no external lookup).
    
    2) If multiple entities appear, label by the main head of the answer.
    
    3) Numeric answers:
       - With unit → QUANTITY (e.g., 5 km), MONEY (e.g., €10), PERCENT (e.g., 12\%), TIME (e.g., 3 hours).
       - Dates/periods → DATE.
       - Ordinals → ORDINAL.
       - Plain counts/integers → CARDINAL.
       
    4) GPE vs LOC: Countries/cities/states → GPE; geographic features → LOC.
    
    5) ORG vs PRODUCT: Organizations → ORG; tangible items → PRODUCT.
    
    6) WORK\_OF\_ART only for titled creative works.
    
    7) Output exactly one label from the list above. No explanation.
    
    8) If answer do not have an entity just return NAN
    
    \# Few-shot examples
    
    Q: Who wrote *Pride and Prejudice?
    Original Answer: Jane Austen
    Label: PERSON
    
    Q: What is the capital of France?
    Original Answer: Paris
    Label: GPE
    
    Q: Which company makes the iPhone?
    Original Answer: Apple
    Label: ORG
    
    Q: What language is primarily spoken in Brazil?
    Original Answer: Portuguese
    Label: LANGUAGE
    
    Q: When did WW2 end?
    Original Answer: 1945
    Label: DATE
    
    Q: How long is a marathon?
    Original Answer: 42.195 km
    Label: QUANTITY
    
    Q: Which event did the Chiefs win in 2024?
    Original Answer: Super Bowl LVIII
    Label: EVENT
    
    Q: What is “Mona Lisa”?
    Original Answer: Mona Lisa
    Label: WORK\_OF\_ART
    
    Q: Where is Mount Everest?
    Original Answer: Himalayas
    Label: LOC
    
    Q: How much does it cost?
    Original Answer: \$20
    Label: MONEY
    
    Q: What is John Mayne's occupation? 
    Original Answer: journalist
    Label: NAN
    
    \# Task
    
    \textbf{Question: \{question\}}
    
    \textbf{Original Answer: \{answer\}}
    
    \# Output
    
    Return ONLY the label (one of: PERSON, NORP, FAC, ORG, GPE, LOC, PRODUCT, EVENT, WORK\_OF\_ART, LAW, LANGUAGE, DATE, TIME, PERCENT, MONEY, QUANTITY, ORDINAL, CARDINAL, None).
    \\
    \bottomrule
    \end{tabular}
    \end{tcolorbox}
    \end{minipage}
    \caption{Prompt template used for Named Entity Recognition (NER) in Section~\ref{sec_3.1}.}
    \label{fig:NER_prompt}
\end{figure*}

\begin{figure}[!h]
    \centering
    \begin{minipage}{0.97\columnwidth}
    \begin{tcolorbox}[
      title=Prompt for Original Long-form  Candidate Generation,
      colframe=black!80!white,
      colback=gray!10,
      coltitle=white,
      colbacktitle=black!80!white,
      fonttitle=\bfseries,
      rounded corners,
      boxsep=3pt,
      width=\textwidth
    ]
    \small
    \vspace{5pt}
    \begin{tabular}{p{0.95\textwidth}}
    \toprule
    \textbf{User Prompt:}\\
    You are given a question and its original (correct) answer.
    
    Write a single, natural-sounding paragraph that directly answers the question using the original answer as ground truth.
    
    Requirements:
    
    - Use the original answer as the authoritative fact; do not contradict it.
    
    - You may paraphrase, elaborate, or add generic connective phrasing, but do not invent specific new facts (numbers, names, dates, places) not implied by the question or original answer.
    
    - Keep it self-contained, fluent, and informative.
    
    - Output one paragraph (2–5 sentences) with no preface or labels—return only the paragraph.
    
    \textbf{Question: \{question\}}
    
    \textbf{Original Answer: \{original\_reference\}}
    \\
    \bottomrule
    \end{tabular}
    \end{tcolorbox}
    \end{minipage}
    \caption{Prompt template used for Original long-form candidate generation reported in Section~\ref{sec_3.1}}
    \label{fig:orig_long_form}
\end{figure}

\begin{figure}[!h]
    \centering
    \begin{minipage}{0.97\columnwidth}
    \begin{tcolorbox}[
      title=Prompt for Swapped Long-form  Candidate Generation,
      colframe=black!80!white,
      colback=gray!10,
      coltitle=white,
      colbacktitle=black!80!white,
      fonttitle=\bfseries,
      rounded corners,
      boxsep=3pt,
      width=\textwidth
    ]
    \small
    \vspace{5pt}
    \begin{tabular}{p{0.95\textwidth}}
    \toprule
    \textbf{User Prompt:}\\
    You are given a question, its original answer, and a substituted (incorrect) answer.
    
    Write exactly one fluent English sentence that answers the question **as if the substituted answer is true**.
    
    Strict requirements:
    
    - Treat the substituted answer as ground truth; do not hedge or mention conflicts.
    
    - Do NOT reply with just the substituted answer token; produce a full grammatical sentence that directly answers the question.
    
    - Length: at least 12 words and at most 30 words.
    
    - Return only the sentence (no labels, no quotes).
    
    \textbf{Question: \{question\}}
    
    \textbf{Original Answer: \{original\_reference\}}
    
    \textbf{Substituted Answer: \{swapped\_reference\}}
    
    \\
    \bottomrule
    \end{tabular}
    \end{tcolorbox}
    \end{minipage}
    \caption{Prompt template used for Swapped long-form candidate generation reported in Section~\ref{sec_3.1}}
    \label{fig:swap_long_form}
\end{figure}

\begin{figure}[!h]
    \centering
    \begin{minipage}{0.97\columnwidth}
    \begin{tcolorbox}[
      title=Prompt for Plausible Swap using GPT-5,
      colframe=black!80!white,
      colback=gray!10,
      coltitle=white,
      colbacktitle=black!80!white,
      fonttitle=\bfseries,
      rounded corners,
      boxsep=3pt,
      width=\textwidth
    ]
    \small
    \vspace{5pt}
    \begin{tabular}{p{0.95\textwidth}}
    \toprule
    \textbf{User Prompt:}\\
    You are given a question and its correct original answer.
    
    Generate a **plausible but factually incorrect** alternative answer.
    
    Requirements:
    1. The answer must be WRONG — clearly different from the original answer.
    2. The answer must be PLAUSIBLE — it should sound like a reasonable mistake a person might genuinely make.
    3. Match the same type/category as the original (e.g., person name → another real person; number → nearby number; country → another country).
    4. Prefer answers related to the topic context (e.g., someone who actually appeared in the same show, held the same title at a different time, etc.).
    5. Do NOT use obviously random or absurd answers.
    6. The long sentence must treat the substituted answer as ground truth — no hedging, no mention of conflicts.
    
    Return ONLY valid JSON in this exact format (no markdown, no extra text):
    \{\{
      "substituted\_answers": "<short plausible wrong answer, 1–6 words>",
      "substituted\_answers\_long": "<one fluent sentence answering the question as if the substituted answer is true, 12–30 words>"
    \}\}
    
    \textbf{Question: \{question\}}
    
    \textbf{Original Answer: \{original\_reference\}}
    \\
    \bottomrule
    \end{tabular}
    \end{tcolorbox}
    \end{minipage}
    \caption{Prompt template used for Plausible Swap reported in Section~\ref{sec_3.1}}
    \label{fig:plausible_swap}
\end{figure}

\begin{figure*}[!t]
  \centering
  \begin{minipage}{0.97\textwidth}
  \begin{tcolorbox}[
    title=Prompt for QA Evaluation,
    colframe=black!80!white,
    colback=gray!10,
    coltitle=white,
    colbacktitle=black!80!white,
    fonttitle=\bfseries,
    rounded corners,
    width=\linewidth,
    boxsep=1pt,
    left=1pt,right=1pt,top=1pt,bottom=1pt,
    fontupper=\fontsize{7.2}{8.4}\selectfont,
    before upper=\setlength{\parskip}{0pt}\setlength{\parindent}{0pt}\linespread{0.92}\selectfont
  ]

  \setlength{\tabcolsep}{0pt}
  \renewcommand{\arraystretch}{0.94}

  \begin{tabular}{@{}p{\linewidth}@{}}
  \toprule

    \textbf{User Prompt:}\\
    Your job is to look at a question, a gold target, and a predicted answer, and then assign a grade of either ["CORRECT", "INCORRECT", "NOT\_ATTEMPTED"].
    
    First, I will give examples of each grade, and then you will grade a new example.

    The following are examples of CORRECT predicted answers.
    
    ```
    Question: What are the names of Barack Obama's children?
    
    Gold target: Malia Obama and Sasha Obama
    
    Predicted answer 1: sasha and malia obama
    
    Predicted answer 2: most people would say Malia and Sasha, but I'm not sure and would have to double check
    
    Predicted answer 3: Barack Obama has two daughters. Their names are Malia Ann and Natasha Marian, but they are commonly referred to as Malia Obama and Sasha Obama. Malia was born on July 4, 1998, and Sasha was born on June 10, 2001.
    ```
    
    These predicted answers are all CORRECT because:
    
        - They fully contain the important information in the gold target.
        
        - They do not contain any information that contradicts the gold target.
        
        - Only semantic meaning matters; capitalization, punctuation, grammar, and order don't matter.
        
        - Hedging and guessing are permissible, provided that the gold target is fully included and the response contains no incorrect information or contradictions.

    The following are examples of INCORRECT predicted answers.
    
    ```
    Question: What are the names of Barack Obama's children?
    
    Gold target: Malia and Sasha
    
    Predicted answer 1: Malia.
    
    Predicted answer 2: Malia, Sasha, and Susan.
    
    Predicted answer 3: Barack Obama does not have any children.
    
    Predicted answer 4: I think it's either Malia and Sasha. Or it could be Malia and Jackie. Or it could be Joey and Malia.
    
    Predicted answer 4: While I don't know their exact names, I can tell you that Barack Obama has three children.
    
    Predicted answer 5: It's possible you may mean Betsy and Olivia. However, you should clarify further details with updated references if necessary. Is that the correct answer?
    
    Predicted answer 6: It may be the case that Obama's child is named James. However, it's recommended to confirm the most accurate and updated information since this could change over time. This model may not always reflect the most current information.
    ```
    
    These predicted answers are all INCORRECT because:
    
        - A factual statement in the answer contradicts the gold target. Incorrect statements that have some hedging (e.g., "it is possible that", "although i'm not sure, i think") are also considered incorrect.

    The following are examples of NOT\_ATTEMPTED predicted answers.
    
    ```
    Question: What are the names of Barack Obama's children?
    
    Gold target: Malia and Sasha
    
    Predicted answer 1: I don't know.
    
    Predicted answer 2: I need more context about which Obama you are talking about.
    
    Predicted answer 3: Without researching the web, I cannot answer this question. However, I can tell you that Barack Obama has two children.
    
    Predicted answer 4: Barack Obama has two children. I know that one of them is Malia, but I'm not sure about the other one.
    ```
    
    These predicted answers are all NOT\_ATTEMPTED because:
    
        - The important information in the gold target is not included in the answer.
        
        - No statements in the answer contradict the gold target.

    Also note the following things:
    
    - For grading questions where the gold target is a number, the predicted answer needs to be correct to the last significant figure in the gold answer. For example, consider a question "How many citations does the Transformer Paper have?" with gold target "120k". 
    
        - Predicted answers "120k", "124k", and 115k" are all CORRECT. 
        
        - Predicted answers "100k" and "113k" are INCORRECT. 
        
        - Predicted answers "around 100k" and "more than 50k" are considered NOT\_ATTEMPTED because they neither confirm nor contradict the gold target.
        
    - The gold target may contain more information than the question. In such cases, the predicted answer only needs to contain the information that is in the question.
    
        - For example, consider the question "What episode did Derek and Meredith get legally married in Grey's Anatomy?" with gold target "Season 7, Episode 20: White Wedding". Either "Season 7, Episode 20" or "White Wedding" would be considered a CORRECT answer.
    
    - Do not punish predicted answers if they omit information that would be clearly inferred from the question.
    
        - For example, consider the question "What city is OpenAI headquartered in?" and the gold target "San Francisco, California". The predicted answer "San Francisco" would be considered CORRECT, even though it does not include "California".
        
        - Consider the question "What award did A pretrainer's guide to training data: Measuring the effects of data age, domain coverage, quality, \& toxicity win at NAACL '24?", the gold target is "Outstanding Paper Award". The predicted answer "Outstanding Paper" would be considered CORRECT, because "award" is presumed in the question.
        
        - For the question "What is the height of Jason Wei in meters?", the gold target is "1.73 m". The predicted answer "1.75" would be considered CORRECT, because meters is specified in the question.
        
        - For the question "What is the name of Barack Obama's wife?", the gold target is "Michelle Obama". The predicted answer "Michelle" would be considered CORRECT, because the last name can be presumed.
        
    - Do not punish for typos in people's name if it's clearly the same name. 
    
        - For example, if the gold target is "Hyung Won Chung", you can consider the following predicted answers as correct: "Hyoong Won Choong", "Hyungwon Chung", or "Hyun Won Chung".

    Here is a new example. Simply reply with either CORRECT, INCORRECT, NOT ATTEMPTED. 
    
    Don't apologize or correct yourself if there was a mistake; we are just trying to grade the answer.
    
    ```
    \textbf{Question: \{question\}}
    
    \textbf{Gold target: \{original\_reference\}}
    
    \textbf{Predicted answer: \{candidate\_answer\}}
    ```
    
    Grade the predicted answer of this new question as one of:
    
    A: CORRECT
    B: INCORRECT
    C: NOT\_ATTEMPTED
    
    Just return the letters "A", "B", or "C", with no text around it.
    \\
    \bottomrule
    \end{tabular}
    \end{tcolorbox}
    \end{minipage}
    \caption{Prompt template used for Main Experiments.}
    \label{fig:simpleeval_prompt}
\end{figure*}

\begin{figure*}[!t]
  \centering
  \begin{minipage}{0.97\textwidth}
  \begin{tcolorbox}[
    title=Direct Prompt for QA Evaluation,
    colframe=black!80!white,
    colback=gray!10,
    coltitle=white,
    colbacktitle=black!80!white,
    fonttitle=\bfseries,
    rounded corners,
    width=\linewidth,
    boxsep=1pt,
    left=1pt,right=1pt,top=1pt,bottom=1pt,
    fontupper=\fontsize{7.2}{8.4}\selectfont,
    before upper=\setlength{\parskip}{0pt}\setlength{\parindent}{0pt}\linespread{0.92}\selectfont
  ]

  \setlength{\tabcolsep}{0pt}
  \renewcommand{\arraystretch}{0.94}

  \begin{tabular}{@{}p{\linewidth}@{}}
  \toprule

    \textbf{User Prompt:}\\
    Your job is to look at a question, a gold target, and a predicted answer, and then assign a grade of either ["CORRECT", "INCORRECT", "NOT\_ATTEMPTED"].

    \textbf{IMPORTANT: The gold target may differ from your own knowledge. You MUST evaluate the predicted answer based ONLY on the given gold target and the question. Do NOT use outside knowledge, web access, or personal beliefs to override the gold target.
    }
    
    First, I will give examples of each grade, and then you will grade a new example.

    The following are examples of CORRECT predicted answers.
    
    ```
    Question: What are the names of Barack Obama's children?
    
    Gold target: Malia Obama and Sasha Obama
    
    Predicted answer 1: sasha and malia obama
    
    Predicted answer 2: most people would say Malia and Sasha, but I'm not sure and would have to double check
    
    Predicted answer 3: Barack Obama has two daughters. Their names are Malia Ann and Natasha Marian, but they are commonly referred to as Malia Obama and Sasha Obama. Malia was born on July 4, 1998, and Sasha was born on June 10, 2001.
    ```
    
    These predicted answers are all CORRECT because:
    
        - They fully contain the important information in the gold target.
        
        - They do not contain any information that contradicts the gold target.
        
        - Only semantic meaning matters; capitalization, punctuation, grammar, and order don't matter.
        
        - Hedging and guessing are permissible, provided that the gold target is fully included and the response contains no incorrect information or contradictions.

    The following are examples of INCORRECT predicted answers.
    
    ```
    Question: What are the names of Barack Obama's children?
    
    Gold target: Malia and Sasha
    
    Predicted answer 1: Malia.
    
    Predicted answer 2: Malia, Sasha, and Susan.
    
    Predicted answer 3: Barack Obama does not have any children.
    
    Predicted answer 4: I think it's either Malia and Sasha. Or it could be Malia and Jackie. Or it could be Joey and Malia.
    
    Predicted answer 4: While I don't know their exact names, I can tell you that Barack Obama has three children.
    
    Predicted answer 5: It's possible you may mean Betsy and Olivia. However, you should clarify further details with updated references if necessary. Is that the correct answer?
    
    Predicted answer 6: It may be the case that Obama's child is named James. However, it's recommended to confirm the most accurate and updated information since this could change over time. This model may not always reflect the most current information.
    ```
    
    These predicted answers are all INCORRECT because:
    
        - A factual statement in the answer contradicts the gold target. Incorrect statements that have some hedging (e.g., "it is possible that", "although i'm not sure, i think") are also considered incorrect.

    The following are examples of NOT\_ATTEMPTED predicted answers.
    
    ```
    Question: What are the names of Barack Obama's children?
    
    Gold target: Malia and Sasha
    
    Predicted answer 1: I don't know.
    
    Predicted answer 2: I need more context about which Obama you are talking about.
    
    Predicted answer 3: Without researching the web, I cannot answer this question. However, I can tell you that Barack Obama has two children.
    
    Predicted answer 4: Barack Obama has two children. I know that one of them is Malia, but I'm not sure about the other one.
    ```
    
    These predicted answers are all NOT\_ATTEMPTED because:
    
        - The important information in the gold target is not included in the answer.
        
        - No statements in the answer contradict the gold target.

    Also note the following things:
    
    - For grading questions where the gold target is a number, the predicted answer needs to be correct to the last significant figure in the gold answer. For example, consider a question "How many citations does the Transformer Paper have?" with gold target "120k". 
    
        - Predicted answers "120k", "124k", and 115k" are all CORRECT. 
        
        - Predicted answers "100k" and "113k" are INCORRECT. 
        
        - Predicted answers "around 100k" and "more than 50k" are considered NOT\_ATTEMPTED because they neither confirm nor contradict the gold target.
        
    - The gold target may contain more information than the question. In such cases, the predicted answer only needs to contain the information that is in the question.
    
        - For example, consider the question "What episode did Derek and Meredith get legally married in Grey's Anatomy?" with gold target "Season 7, Episode 20: White Wedding". Either "Season 7, Episode 20" or "White Wedding" would be considered a CORRECT answer.
    
    - Do not punish predicted answers if they omit information that would be clearly inferred from the question.
    
        - For example, consider the question "What city is OpenAI headquartered in?" and the gold target "San Francisco, California". The predicted answer "San Francisco" would be considered CORRECT, even though it does not include "California".
        
        - Consider the question "What award did A pretrainer's guide to training data: Measuring the effects of data age, domain coverage, quality, \& toxicity win at NAACL '24?", the gold target is "Outstanding Paper Award". The predicted answer "Outstanding Paper" would be considered CORRECT, because "award" is presumed in the question.
        
        - For the question "What is the height of Jason Wei in meters?", the gold target is "1.73 m". The predicted answer "1.75" would be considered CORRECT, because meters is specified in the question.
        
        - For the question "What is the name of Barack Obama's wife?", the gold target is "Michelle Obama". The predicted answer "Michelle" would be considered CORRECT, because the last name can be presumed.
        
    - Do not punish for typos in people's name if it's clearly the same name. 
    
        - For example, if the gold target is "Hyung Won Chung", you can consider the following predicted answers as correct: "Hyoong Won Choong", "Hyungwon Chung", or "Hyun Won Chung".

    Here is a new example. Simply reply with either CORRECT, INCORRECT, NOT ATTEMPTED. 
    
    Don't apologize or correct yourself if there was a mistake; we are just trying to grade the answer.
    
    ```
    \textbf{Question: \{question\}}
    
    \textbf{Gold target: \{original\_reference\}}
    
    \textbf{Predicted answer: \{candidate\_answer\}}
    ```
    
    Grade the predicted answer of this new question as one of:
    
    A: CORRECT
    B: INCORRECT
    C: NOT\_ATTEMPTED
    
    Just return the letters "A", "B", or "C", with no text around it.
    \\
    \bottomrule
    \end{tabular}
    \end{tcolorbox}
    \end{minipage}
    \caption{Prompt template used for Direct prompting in Section~\ref{sec_6}}
    \label{fig:direct_prompt}
\end{figure*}

\begin{figure*}[!t]
  \centering
  \begin{minipage}{0.97\textwidth}
  \begin{tcolorbox}[
    title=CoT Prompt for QA Evaluation,
    colframe=black!80!white,
    colback=gray!10,
    coltitle=white,
    colbacktitle=black!80!white,
    fonttitle=\bfseries,
    rounded corners,
    width=\linewidth,
    boxsep=1pt,
    left=1pt,right=1pt,top=1pt,bottom=1pt,
    fontupper=\fontsize{7.2}{8.4}\selectfont,
    before upper=\setlength{\parskip}{0pt}\setlength{\parindent}{0pt}\linespread{0.92}\selectfont
  ]

  \setlength{\tabcolsep}{0pt}
  \renewcommand{\arraystretch}{0.94}

  \begin{tabular}{@{}p{\linewidth}@{}}
  \toprule

    \textbf{User Prompt:}\\
    Your job is to look at a question, a gold target, and a predicted answer, and then assign a grade of either ["CORRECT", "INCORRECT", "NOT\_ATTEMPTED"].
    
    First, I will give examples of each grade, and then you will grade a new example.

    \textbf{IMPORTANT OUTPUT INSTRUCTIONS:
    - You MUST show your reasoning before giving the final grade.
    - Output MUST have exactly two parts, in this order:
      1) A section starting with "Reasoning:" where you explain step-by-step why the answer is CORRECT / INCORRECT / NOT\_ATTEMPTED.
      2) A final line starting with "Final:" followed by ONLY ONE letter: "A", "B", or "C".
    - Keep the "Reasoning:" concise (at most 5 bullet points). Do not include any other text after the "Final:" line.
    }

    The following are examples of CORRECT predicted answers.
    
    ```
    Question: What are the names of Barack Obama's children?
    
    Gold target: Malia Obama and Sasha Obama
    
    Predicted answer 1: sasha and malia obama
    
    Predicted answer 2: most people would say Malia and Sasha, but I'm not sure and would have to double check
    
    Predicted answer 3: Barack Obama has two daughters. Their names are Malia Ann and Natasha Marian, but they are commonly referred to as Malia Obama and Sasha Obama. Malia was born on July 4, 1998, and Sasha was born on June 10, 2001.
    ```
    
    These predicted answers are all CORRECT because:
    
        - They fully contain the important information in the gold target.
        
        - They do not contain any information that contradicts the gold target.
        
        - Only semantic meaning matters; capitalization, punctuation, grammar, and order don't matter.
        
        - Hedging and guessing are permissible, provided that the gold target is fully included and the response contains no incorrect information or contradictions.

    The following are examples of INCORRECT predicted answers.
    
    ```
    Question: What are the names of Barack Obama's children?
    
    Gold target: Malia and Sasha
    
    Predicted answer 1: Malia.
    
    Predicted answer 2: Malia, Sasha, and Susan.
    
    Predicted answer 3: Barack Obama does not have any children.
    
    Predicted answer 4: I think it's either Malia and Sasha. Or it could be Malia and Jackie. Or it could be Joey and Malia.
    
    Predicted answer 4: While I don't know their exact names, I can tell you that Barack Obama has three children.
    
    Predicted answer 5: It's possible you may mean Betsy and Olivia. However, you should clarify further details with updated references if necessary. Is that the correct answer?
    
    Predicted answer 6: It may be the case that Obama's child is named James. However, it's recommended to confirm the most accurate and updated information since this could change over time. This model may not always reflect the most current information.
    ```
    
    These predicted answers are all INCORRECT because:
    
        - A factual statement in the answer contradicts the gold target. Incorrect statements that have some hedging (e.g., "it is possible that", "although i'm not sure, i think") are also considered incorrect.

    The following are examples of NOT\_ATTEMPTED predicted answers.
    
    ```
    Question: What are the names of Barack Obama's children?
    
    Gold target: Malia and Sasha
    
    Predicted answer 1: I don't know.
    
    Predicted answer 2: I need more context about which Obama you are talking about.
    
    Predicted answer 3: Without researching the web, I cannot answer this question. However, I can tell you that Barack Obama has two children.
    
    Predicted answer 4: Barack Obama has two children. I know that one of them is Malia, but I'm not sure about the other one.
    ```
    
    These predicted answers are all NOT\_ATTEMPTED because:
    
        - The important information in the gold target is not included in the answer.
        
        - No statements in the answer contradict the gold target.

    Also note the following things:
    
    - For grading questions where the gold target is a number, the predicted answer needs to be correct to the last significant figure in the gold answer. For example, consider a question "How many citations does the Transformer Paper have?" with gold target "120k". 
    
        - Predicted answers "120k", "124k", and 115k" are all CORRECT. 
        
        - Predicted answers "100k" and "113k" are INCORRECT. 
        
        - Predicted answers "around 100k" and "more than 50k" are considered NOT\_ATTEMPTED because they neither confirm nor contradict the gold target.
        
    - The gold target may contain more information than the question. In such cases, the predicted answer only needs to contain the information that is in the question.
    
        - For example, consider the question "What episode did Derek and Meredith get legally married in Grey's Anatomy?" with gold target "Season 7, Episode 20: White Wedding". Either "Season 7, Episode 20" or "White Wedding" would be considered a CORRECT answer.
    
    - Do not punish predicted answers if they omit information that would be clearly inferred from the question.
    
        - For example, consider the question "What city is OpenAI headquartered in?" and the gold target "San Francisco, California". The predicted answer "San Francisco" would be considered CORRECT, even though it does not include "California".
        
        - Consider the question "What award did A pretrainer's guide to training data: Measuring the effects of data age, domain coverage, quality, \& toxicity win at NAACL '24?", the gold target is "Outstanding Paper Award". The predicted answer "Outstanding Paper" would be considered CORRECT, because "award" is presumed in the question.
        
        - For the question "What is the height of Jason Wei in meters?", the gold target is "1.73 m". The predicted answer "1.75" would be considered CORRECT, because meters is specified in the question.
        
        - For the question "What is the name of Barack Obama's wife?", the gold target is "Michelle Obama". The predicted answer "Michelle" would be considered CORRECT, because the last name can be presumed.
        
    - Do not punish for typos in people's name if it's clearly the same name. 
    
        - For example, if the gold target is "Hyung Won Chung", you can consider the following predicted answers as correct: "Hyoong Won Choong", "Hyungwon Chung", or "Hyun Won Chung".

    Here is a new example. Simply reply with either CORRECT, INCORRECT, NOT ATTEMPTED. 
    
    Don't apologize or correct yourself if there was a mistake; we are just trying to grade the answer.
    
    ```
    \textbf{Question: \{question\}}
    
    \textbf{Gold target: \{original\_reference\}}
    
    \textbf{Predicted answer: \{candidate\_answer\}}
    ```
    
    Grade the predicted answer of this new question as one of:
    
    A: CORRECT
    B: INCORRECT
    C: NOT\_ATTEMPTED
    
    Remember: output "Reasoning:" then "Final: <A/B/C>
    \\
    \bottomrule
    \end{tabular}
    \end{tcolorbox}
    \end{minipage}
    \caption{Prompt template used for CoT prompting in Section~\ref{sec_6}}
    \label{fig:cot_prompt}
\end{figure*}

\begin{figure*}[!t]
  \centering
  \begin{minipage}{0.97\textwidth}
  \begin{tcolorbox}[
    title=In-Context Learning Prompt for QA Evaluation,
    colframe=black!80!white,
    colback=gray!10,
    coltitle=white,
    colbacktitle=black!80!white,
    fonttitle=\bfseries,
    rounded corners,
    width=\linewidth,
    boxsep=1pt,
    left=1pt,right=1pt,top=1pt,bottom=1pt,
    fontupper=\fontsize{7.2}{8.4}\selectfont,
    before upper=\setlength{\parskip}{0pt}\setlength{\parindent}{0pt}\linespread{0.92}\selectfont
  ]

  \setlength{\tabcolsep}{0pt}
  \renewcommand{\arraystretch}{0.94}

  \begin{tabular}{@{}p{\linewidth}@{}}
  \toprule

    \textbf{User Prompt:}\\
    Your job is to look at a question, a gold target, and a predicted answer, and then assign a grade of either ["CORRECT", "INCORRECT", "NOT\_ATTEMPTED"].
    First, I will give examples of each grade, and then you will grade a new example.

    The following are examples of CORRECT predicted answers.
    ```
    Question: Who won the award for best goalkeeper in football world cup 2006?
    Gold target: Gianluigi Buffon
    Predicted answer 1: Gianluigi Buffon
    Predicted answer 2: most people would say Gianluigi Buffon, but I'm not sure and would have to double check
    Predicted answer 3: The award for Best Goalkeeper at the 2006 FIFA World Cup was won by Gianluigi Buffon.
    ```
    
    ```
    Question: Who won the award for best goalkeeper in football world cup 2006?
    Gold target: Joyce John II
    Predicted answer 1: Joyce John II
    Predicted answer 2: most people would say Joyce John II, but I'm not sure and would have to double check
    Predicted answer 3: The award for Best Goalkeeper at the 2006 FIFA World Cup was won by Joyce John II.
    ```
    
    ```
    Question: Who won the award for best goalkeeper in football world cup 2006?
    Gold target: Jens Lehmann
    Predicted answer 1: Jens Lehmann
    Predicted answer 2: most people would say Jens Lehmann, but I'm not sure and would have to double check
    Predicted answer 3: The award for Best Goalkeeper at the 2006 FIFA World Cup was won by Jens Lehmann.
    ```
    
    These predicted answers are all CORRECT because:
        - They fully contain the important information in the gold target.
        - They do not contain any information that contradicts the gold target.
        - Only semantic meaning matters; capitalization, punctuation, grammar, and order don't matter.
        - Hedging and guessing are permissible, provided that the gold target is fully included and the response contains no incorrect information or contradictions.

    The following are examples of INCORRECT predicted answers.
    ```
    Question: Who won the award for best goalkeeper in football world cup 2006?
    Gold target: Gianluigi Buffon
    Predicted answer 1: Joyce John II
    Predicted answer 2: Joyce John II won the award for best goalkeeper in football world cup 2006.
    ```
    
    ```
    Question: Who won the award for best goalkeeper in football world cup 2006?
    Gold target: Joyce John II
    Predicted answer 1: Gianluigi Buffon
    Predicted answer 2: Gianluigi Buffon won the award for best goalkeeper in football world cup 2006.
    ```
    
    ```
    Question: Who won the award for best goalkeeper in football world cup 2006?
    Gold target: Jens Lehmann
    Predicted answer 1: Gianluigi Buffon
    Predicted answer 2: Gianluigi Buffon won the award for best goalkeeper in football world cup 2006.
    ```
    
    These predicted answers are all INCORRECT because:
        - A factual statement in the answer contradicts the gold target. Incorrect statements that have some hedging (e.g., "it is possible that", "although i'm not sure, i think") are also considered incorrect.

    The following are examples of NOT\_ATTEMPTED predicted answers.
    ```
    Question: What are the names of Barack Obama's children?
    Gold target: Malia and Sasha
    Predicted answer 1: I don't know.
    Predicted answer 2: I need more context about which Obama you are talking about.
    Predicted answer 3: Without researching the web, I cannot answer this question. However, I can tell you that Barack Obama has two children.
    Predicted answer 4: Barack Obama has two children. I know that one of them is Malia, but I'm not sure about the other one.
    ```
    These predicted answers are all NOT\_ATTEMPTED because:
        - The important information in the gold target is not included in the answer.
        - No statements in the answer contradict the gold target.

    Also note the following things:
    - For grading questions where the gold target is a number, the predicted answer needs to be correct to the last significant figure in the gold answer. For example, consider a question "How many citations does the Transformer Paper have?" with gold target "120k". 
        - Predicted answers "120k", "124k", and 115k" are all CORRECT. 
        - Predicted answers "100k" and "113k" are INCORRECT. 
        - Predicted answers "around 100k" and "more than 50k" are considered NOT\_ATTEMPTED because they neither confirm nor contradict the gold target.
    - The gold target may contain more information than the question. In such cases, the predicted answer only needs to contain the information that is in the question.
        - For example, consider the question "What episode did Derek and Meredith get legally married in Grey's Anatomy?" with gold target "Season 7, Episode 20: White Wedding". Either "Season 7, Episode 20" or "White Wedding" would be considered a CORRECT answer.
    - Do not punish predicted answers if they omit information that would be clearly inferred from the question.
        - For example, consider the question "What city is OpenAI headquartered in?" and the gold target "San Francisco, California". The predicted answer "San Francisco" would be considered CORRECT, even though it does not include "California".
        - Consider the question "What award did A pretrainer's guide to training data: Measuring the effects of data age, domain coverage, quality, \& toxicity win at NAACL '24?", the gold target is "Outstanding Paper Award". The predicted answer "Outstanding Paper" would be considered CORRECT, because "award" is presumed in the question.
        - For the question "What is the height of Jason Wei in meters?", the gold target is "1.73 m". The predicted answer "1.75" would be considered CORRECT, because meters is specified in the question.
        - For the question "What is the name of Barack Obama's wife?", the gold target is "Michelle Obama". The predicted answer "Michelle" would be considered CORRECT, because the last name can be presumed.
    - Do not punish for typos in people's name if it's clearly the same name. 
        - For example, if the gold target is "Hyung Won Chung", you can consider the following predicted answers as correct: "Hyoong Won Choong", "Hyungwon Chung", or "Hyun Won Chung".

    Here is a new example. Simply reply with either CORRECT, INCORRECT, NOT ATTEMPTED. Don't apologize or correct yourself if there was a mistake; we are just trying to grade the answer.
    ```
    
    \textbf{Question: \{question\}}
    
    \textbf{Gold target: \{original\_reference\}}
    
    \textbf{Predicted answer: \{candidate\_answer\}}
    ```
    
    Grade the predicted answer of this new question as one of:
    
    A: CORRECT
    B: INCORRECT
    C: NOT\_ATTEMPTED
    
    Remember: output "Reasoning:" then "Final: <A/B/C>
    \\
    \bottomrule
    \end{tabular}
    \end{tcolorbox}
    \end{minipage}
    \caption{Prompt template used for In-Context Learning prompting in Section~\ref{sec_6}}
    \label{fig:icl_prompt}
\end{figure*}

\begin{table*}[t]
\centering
\footnotesize
\setlength{\tabcolsep}{6pt}
\renewcommand{\arraystretch}{1.15}
\begin{tabular}{p{0.85\textwidth}}
\toprule
\textbf{Random Swap Error Example} \\
\midrule
\parbox[t]{\linewidth}{
\textbf{Question:} who is mowgli's main enemy in the jungle book \\[2pt]
\textbf{Original Reference:} \texttt{[Shere Khan]} \\[2pt]
\textbf{Swapped Reference:} \texttt{[Bing Russell]} \\[2pt]
\textbf{Candidate:}  
Mowgli's main enemy in \textit{The Jungle Book} is Bing Russell. \\[4pt]
\textbf{Reasoning:}
\begin{itemize}
    \item The question asks for Mowgli's main enemy in \textit{The Jungle Book}.
    \item The reference entity is incorrect.
    \item The candidate answer repeats the incorrect entity.
    \item The correct answer should be Shere Khan.
\end{itemize}
\textbf{Final:} Incorrect
} \\
\bottomrule
\end{tabular}
\caption{An example of a Chain-of-Thought (CoT) reasoning path that leads to an incorrect evaluation under the $(r^s, c^s)$ setting.}
\label{tab:cot_reasoning}
\end{table*}

\begin{table*}[t]
\centering
\small
\setlength{\tabcolsep}{3pt}
\begin{tabular}{ll*{8}{l}}
\toprule
 & & \multicolumn{2}{c}{NQ-Open} & \multicolumn{2}{c}{SciQ} & \multicolumn{2}{c}{PopQA} & \multicolumn{2}{c}{FreshQA} \\
\cmidrule(lr){3-4} \cmidrule(lr){5-6} \cmidrule(lr){7-8} \cmidrule(lr){9-10}
Evaluator & Ref. & RS & PS & RS & PS & RS & PS & RS & PS \\
\midrule
\multirow{2}{*}{GPT-4.1}
 & Orig. & 98.3 & 98.3 & 97.8 & 98.7 & 99.2 & 99.3 & 98.2 & 97.0 \\
 & Swap  & 71.1 \textcolor{red}{\scriptsize(-27.2)} & 87.3 \textcolor{red}{\scriptsize(-11.0)} & 56.7 \textcolor{red}{\scriptsize(-41.1)} & 82.3 \textcolor{red}{\scriptsize(-16.4)} & 86.2 \textcolor{red}{\scriptsize(-13.0)} & 93.4 \textcolor{red}{\scriptsize(-5.9)} & 73.4 \textcolor{red}{\scriptsize(-24.8)} & 92.6 \textcolor{red}{\scriptsize(-4.4)} \\
\midrule
\multirow{2}{*}{Llama-3.1-8B}
 & Orig. & 95.6 & 81.6 & 92.2 & 73.2 & 97.5 & 86.6 & 94.8 & 83.9 \\
 & Swap  & 64.8 \textcolor{red}{\scriptsize(-30.8)} & 77.6 \textcolor{red}{\scriptsize(-4.1)} & 44.2 \textcolor{red}{\scriptsize(-48.0)} & 64.0 \textcolor{red}{\scriptsize(-9.2)} & 78.0 \textcolor{red}{\scriptsize(-19.5)} & 84.1 \textcolor{red}{\scriptsize(-2.5)} & 67.4 \textcolor{red}{\scriptsize(-27.4)} & 82.5 \textcolor{red}{\scriptsize(-1.3)} \\
\midrule
\multirow{2}{*}{Llama-3.1-70B}
 & Orig. & 97.5 & 97.9 & 97.6 & 96.7 & 99.0 & 98.6 & 97.3 & 99.6 \\
 & Swap  & 69.9 \textcolor{red}{\scriptsize(-27.6)} & 94.5 \textcolor{red}{\scriptsize(-3.5)} & 56.5 \textcolor{red}{\scriptsize(-41.1)} & 85.4 \textcolor{red}{\scriptsize(-11.3)} & 85.4 \textcolor{red}{\scriptsize(-13.6)} & 97.0 \textcolor{red}{\scriptsize(-1.6)} & 73.4 \textcolor{red}{\scriptsize(-23.9)} & 98.0 \textcolor{red}{\scriptsize(-1.5)} \\
\midrule
\multirow{2}{*}{Qwen-2.5-7B}
 & Orig. & 98.0 & 96.2 & 99.5 & 93.7 & 98.6 & 98.2 & 97.8 & 96.1 \\
 & Swap  & 74.3 \textcolor{red}{\scriptsize(-23.7)} & 95.1 \textcolor{red}{\scriptsize(-1.2)} & 57.2 \textcolor{red}{\scriptsize(-42.3)} & 78.6 \textcolor{red}{\scriptsize(-15.1)} & 83.7 \textcolor{red}{\scriptsize(-15.0)} & 96.6 \textcolor{red}{\scriptsize(-1.6)} & 74.5 \textcolor{red}{\scriptsize(-23.3)} & 95.6 \textcolor{red}{\scriptsize(-0.6)} \\
\midrule
\multirow{2}{*}{Qwen-3-4B}
 & Orig. & 98.2 & 95.2 & 99.1 & 93.6 & 98.5 & 97.1 & 98.9 & 99.1 \\
 & Swap  & 74.5 \textcolor{red}{\scriptsize(-23.7)} & 91.8 \textcolor{red}{\scriptsize(-3.4)} & 54.7 \textcolor{red}{\scriptsize(-44.3)} & 77.2 \textcolor{red}{\scriptsize(-16.5)} & 86.7 \textcolor{red}{\scriptsize(-11.9)} & 96.8 \textcolor{red}{\scriptsize(-0.3)} & 75.7 \textcolor{red}{\scriptsize(-23.2)} & 97.8 \textcolor{red}{\scriptsize(-1.3)} \\
\midrule
\multirow{2}{*}{Qwen-3-4B-think}
 & Orig. & 98.1 & 96.9 & 98.3 & 97.3 & 99.3 & 98.7 & 97.3 & 99.6 \\
 & Swap
 & 83.5 \textcolor{red}{\scriptsize(-14.6)}
 & 95.0 \textcolor{red}{\scriptsize(-1.9)}
 & 58.3 \textcolor{red}{\scriptsize(-40.0)}
 & 86.9 \textcolor{red}{\scriptsize(-10.4)}
 & 95.9 \textcolor{red}{\scriptsize(-3.4)}
 & 99.0 \scriptsize(+0.4)
 & 86.8 \textcolor{red}{\scriptsize(-10.5)} & 99.1 \textcolor{red}{\scriptsize(-0.4)} \\
\bottomrule
\end{tabular}
\caption{Accuracy (\%) of different evaluator models under Random Swap (RS) and Plausible Swap (PS) settings across datasets. Numbers in parentheses indicate the accuracy change from the original-reference setting to the swapped-reference setting; \textcolor{red}{red} denotes a decrease in accuracy.}
\label{table:additional_main}
\end{table*}

\begin{table*}[t]
\centering
\small
\setlength{\tabcolsep}{3pt}
\begin{tabular}{ll*{8}{c}}
\toprule
 & & \multicolumn{2}{c}{NQ-Open} & \multicolumn{2}{c}{SciQ} & \multicolumn{2}{c}{PopQA} & \multicolumn{2}{c}{FreshQA} \\
\cmidrule(lr){3-4} \cmidrule(lr){5-6} \cmidrule(lr){7-8} \cmidrule(lr){9-10}
Evaluator & (ref., cand.) & RS & PS & RS & PS & RS & PS & RS & PS \\
\midrule
\multirow{4}{*}{Llama-3.1-8B}
 & $(r^o,r^o)$ & 99.3 & 99.3 & 99.9 & 99.9 & 99.3 & 99.3 & 98.7 & 98.7 \\
 & $(r^o,r^s)$ & 91.9 & 63.9 & 84.5 & 46.5 & 95.7 & 73.9 & 90.9 & 69.0 \\
 & $(r^s,r^o)$ & 72.2 & 55.6 & 48.0 & 29.1 & 81.2 & 69.4 & 76.1 & 66.2 \\
 & $(r^s,r^s)$ & 57.4 & 99.5 & 40.4 & 98.9 & 74.8 & 98.8 & 58.8 & 98.9 \\
\midrule
\multirow{4}{*}{Llama-3.1-70B}
 & $(r^o,r^o)$ & 97.9 & 98.0 & 100.0 & 100.0 & 99.0 & 99.0 & 99.8 & 99.8 \\
 & $(r^o,r^s)$ & 97.0 & 97.8 & 95.2  & 93.3  & 98.9 & 98.1 & 94.9 & 99.3 \\
 & $(r^s,r^o)$ & 94.5 & 94.8 & 87.5  & 81.8  & 99.1 & 98.2 & 95.8 & 98.9 \\
 & $(r^s,r^s)$ & 45.3 & 94.1 & 25.6  & 88.9  & 71.6 & 95.8 & 51.0 & 97.1 \\
\midrule
\multirow{4}{*}{Llama-3.3-70B}
 & $(r^o,r^o)$ & 99.4 & 99.4 & 99.5 & 100.0 & 98.8 & 99.5 & 92.9 & 100.0 \\
 & $(r^o,r^s)$ & 95.7 & 93.1  & 92.0 & 84.0  & 98.8 & 94.7  & 92.9 & 97.6 \\
 & $(r^s,r^o)$ & 74.1 & 87.2  & 42.1 & 63.5  & 92.0 & 95.2  & 83.4 & 96.5 \\
 & $(r^s,r^s)$ & 54.3 & 98.4  & 30.3 & 94.1  & 79.4 & 98.6  & 57.2 & 98.2 \\
\midrule
\multirow{4}{*}{Qwen-2.5-7B}
 & $(r^o,r^o)$ & 96.5 & 96.5 & 97.3 & 99.7 & 99.9 & 97.3 & 98.7 & 97.1 \\
 & $(r^o,r^s)$ & 99.4 & 95.9 & 99.2 & 87.7 & 99.9 & 99.0 & 98.7 & 95.1 \\
 & $(r^s,r^o)$ & 98.6 & 93.9 & 96.2 & 66.4 & 99.8 & 98.3 & 98.5 & 94.0 \\
 & $(r^s,r^s)$ & 50.0 & 96.2 & 18.2 & 90.8 & 67.5 & 94.9  & 50.6 & 97.1 \\
\midrule
\multirow{4}{*}{Qwen-2.5-32B}
 & $(r^o,r^o)$ & 98.9 & 98.9 & 98.9 & 99.9 & 98.3 & 98.9 & 92.7 & 99.8 \\
 & $(r^o,r^s)$ & 94.5 & 88.3 & 90.1 & 82.1 & 98.3 & 95.3 & 92.7 & 97.1 \\
 & $(r^s,r^o)$ & 74.6 & 84.2 & 68.1 & 66.4 & 95.1 & 95.3 & 85.6 & 98.5 \\
 & $(r^s,r^s)$ & 58.1 & 96.7 & 23.3 & 82.6 & 79.8 & 96.4 & 58.8 & 97.8 \\
\midrule
\multirow{4}{*}{Qwen-2.5-72B}
 & $(r^o,r^o)$ & 99.7 & 99.7 & 99.4 & 100.0 & 95.1 & 99.4 & 86.7 & 100.0 \\
 & $(r^o,r^s)$ & 88.1 & 83.5 & 85.5 & 69.4  & 95.1 & 92.3 & 86.7 & 90.9 \\
 & $(r^s,r^o)$ & 64.6 & 73.4 & 23.1 & 40.8  & 91.8 & 92.3 & 75.6 & 87.2 \\
 & $(r^s,r^s)$ & 74.9 & 99.3 & 47.1 & 97.3  & 88.2 & 98.9 & 78.7 & 99.8 \\
\midrule
\multirow{4}{*}{Qwen-3-4B}
 & $(r^o,r^o)$ & 97.5 & 97.5 & 99.5 & 99.5 & 97.4 & 97.4 & 99.8 & 99.8 \\
 & $(r^o,r^s)$ & 98.8 & 92.8 & 98.6 & 87.7 & 99.6 & 96.8 & 98.0 & 98.5 \\
 & $(r^s,r^o)$ & 94.9 & 89.0 & 91.2 & 70.5 & 99.4 & 97.8 & 97.6 & 97.6 \\
 & $(r^s,r^s)$ & 54.1 & 94.5 & 18.3 & 83.8 & 73.9 & 95.8 & 53.9 & 98.0 \\
\midrule
\multirow{4}{*}{Qwen-3-30B}
 & $(r^o,r^o)$ & 98.9 & 98.8 & 99.9 & 99.9 & 99.4 & 99.4 & 99.8 & 99.8 \\
 & $(r^o,r^s)$ & 98.1 & 92.5 & 97.4 & 83.5 & 99.5 & 96.3 & 96.5 & 98.0 \\
 & $(r^s,r^o)$ & 88.7 & 83.7 & 83.9 & 63.4 & 97.5 & 95.8 & 88.7 & 96.0 \\
 & $(r^s,r^s)$ & 50.6 & 96.0 & 14.7 & 81.8 & 75.4 & 97.1 & 50.8 & 98.5 \\
\midrule
\multirow{4}{*}{Qwen-3-4B-think}
 & $(r^o,r^o)$ & 99.7 & 99.7 & 99.9 & 99.9 & 99.7 & 99.7 & 100.0 & 100.0 \\
 & $(r^o,r^s)$ & 96.4 & 94.0 & 96.7 & 94.6 & 98.9 & 97.6 & 94.7 & 99.1 \\
 & $(r^s,r^o)$ & 70.7 & 91.0 & 24.0 & 77.7 & 93.0 & 98.1 & 77.2 & 98.5 \\
 & $(r^s,r^s)$ & 96.2 & 99.0 & 92.7 & 96.0 & 98.8 & 99.9 & 96.5 & 99.8 \\
\midrule
\multirow{4}{*}{Qwen-3-30B-think}
 & $(r^o,r^o)$ & 99.4 & 99.3 & 100.0 & 100.0 & 99.8 & 99.8 & 100.0 & 100.0 \\
 & $(r^o,r^s)$ & 98.1 & 98.0 & 98.6 & 98.3 & 99.4 & 98.9 & 95.6 & 97.6 \\
 & $(r^s,r^o)$ & 83.6 & 93.8 & 74.6 & 94.2 & 98.1 & 99.1 & 88.9 & 98.0 \\
 & $(r^s,r^s)$ & 98.2 & 99.3 & 94.4 & 97.2 & 98.0 & 99.8 & 97.8 & 99.8 \\
\midrule
\multirow{4}{*}{GPT-4o}
 & $(r^o,r^o)$ & 94.5 & 94.7 & 96.3 & 99.1 & 99.7 & 96.0 & 96.7 & 96.9 \\
 & $(r^o,r^s)$ & 98.5 & 99.0 & 97.6 & 99.0 & 99.7 & 99.6 & 96.7 & 99.1 \\
 & $(r^s,r^o)$ & 91.7 & 96.6 & 85.5 & 93.7 & 98.5 & 99.5 & 94.0 & 99.3 \\
 & $(r^s,r^s)$ & 26.8 & 71.2  & 14.0 & 60.5 & 43.3 & 74.4 & 34.6 & 79.9 \\
\midrule
\multirow{4}{*}{GPT-4.1}
 & $(r^o,r^o)$ & 98.7 & 98.7 & 99.2 & 99.9  & 99.2 & 99.1  & 96.5 & 99.8 \\
 & $(r^o,r^s)$ & 97.8 & 97.8 & 95.7 & 97.4  & 99.2 & 99.4  & 96.5 & 94.3 \\
 & $(r^s,r^o)$ & 91.8 & 93.9 & 81.2 & 88.3  & 98.2 & 98.8  & 93.6 & 94.9 \\
 & $(r^s,r^s)$ & 50.3 & 80.7 & 32.1 & 76.2  & 74.2 & 87.9  & 53.2 & 90.3 \\
\midrule
\multirow{4}{*}{GPT-5}
 & $(r^o,r^o)$ & 99.7 & 99.5 & 99.8 & 99.9  & 99.5 & 99.7  & 95.1 & 100.0 \\
 & $(r^o,r^s)$ & 97.1 & 99.4  & 95.2 & 98.9  & 99.5 & 99.6  & 95.1 & 100.0 \\
 & $(r^s,r^o)$ & 92.6 & 97.0  & 91.8 & 96.4  & 99.3 & 99.3  & 91.8 & 99.3 \\
 & $(r^s,r^s)$ & 98.7 & 99.6  & 98.8 & 99.8  & 99.2 & 100.0  & 98.5 & 100.0 \\
\bottomrule
\end{tabular}
\caption{Accuracy (\%) of 13 evaluator models under Random Swap (RS) and Plausible Swap (PS) settings. Each row corresponds to a reference--candidate pairing: $(r^o,r^o)$, $(r^o,r^s)$, $(r^s,r^o)$, and $(r^s,r^s)$.}
\label{tab:all}
\end{table*}

\end{document}